\documentclass{article}

\PassOptionsToPackage{numbers}{natbib}
\usepackage[preprint]{neurips_2026}
\usepackage[utf8]{inputenc} 
\usepackage[T1]{fontenc}    
\usepackage{url}            
\usepackage{booktabs}       
\usepackage{amsfonts}       
\usepackage{nicefrac}       
\usepackage{microtype}      
\usepackage{xcolor}         
\usepackage{makecell}
\usepackage[table]{xcolor}
\usepackage{colortbl}
\usepackage{adjustbox}
\usepackage{xspace}
\usepackage{xltabular}
\usepackage{multirow}
\usepackage{graphicx}
\usepackage{subcaption}
\usepackage{wrapfig}
\usepackage{amsmath}
\usepackage{amssymb}
\definecolor{linkColor}{rgb}{0.18,0.39,0.62}
\usepackage[colorlinks=true,linkcolor=linkColor,citecolor=linkColor,filecolor=linkColor,urlcolor=linkColor]{hyperref}
\newcommand\ourmethod{\textsc{EnvRL}}

\title{\ourmethod{}: Learn from Environment Dynamics in Agentic Reinforcement Learning}

\author{
  \textbf{Zhitong Wang}\thanks{Equal contribution.}\textsuperscript{$\ast\spadesuit$},
  \textbf{Songze Li}\textsuperscript{$\ast\heartsuit$},
  \textbf{Hao Peng}\textsuperscript{$\spadesuit$},
  \textbf{Shuzheng Si}\textsuperscript{$\spadesuit$},
\\
  \textbf{Yi Wang}\textsuperscript{$\heartsuit$},
  \textbf{Maosong Sun}\textsuperscript{$\spadesuit$},
  \textbf{Juanzi Li}\textsuperscript{$\spadesuit$}
\\
  \textsuperscript{$\spadesuit$}Department of Computer Science and Technology, Tsinghua University
\\
  \textsuperscript{$\heartsuit$}Shanghai AI Laboratory 
\\
  \url{https://github.com/zt-wang19/EnvRL}
}

\begin{document}
\maketitle

\begin{abstract}

Reinforcement learning (RL) has emerged as a powerful paradigm for training Large Language Models (LLMs) as agents. 
However, conventional RL methods for long-horizon agentic tasks often struggle with sparse outcome rewards. Intuitively, this overlooks the rich \textbf{environment dynamics} information contained in rollout interaction trajectories. We argue that the interaction experience inherently serves as an implicit supervision signal, reveals the underlying environmental transition mechanisms, and enables the agent to construct a more accurate internal model of the environment.
Therefore, in this work, we investigate how to leverage this additional signal to improve policy learning. Specifically, we propose \textbf{\ourmethod{}}, a framework that incorporates environment dynamics learning into agentic RL via two auxiliary objectives: \textit{state prediction} and \textit{inverse dynamics}. By jointly optimizing with the primary RL objective, we encourage the agent to internalize environment dynamics from its own interaction, leading to more effective decision-making. Extensive experiments on two long-horizon agentic benchmarks demonstrate that \ourmethod{} achieves significant improvements on success-rates over RL-only baselines, e.g., when trained with GRPO, lifting Qwen-2.5-1.5B-Instruct from 72.8\% to 77.4\% on ALFWorld, and from 56.8\% to 67.0\% on WebShop.
\end{abstract}


\section{Introduction}

\begin{wrapfigure}{r}{0.52\textwidth}
    \vspace{-9.0pt}
    \centering
\includegraphics[width=0.52\textwidth]{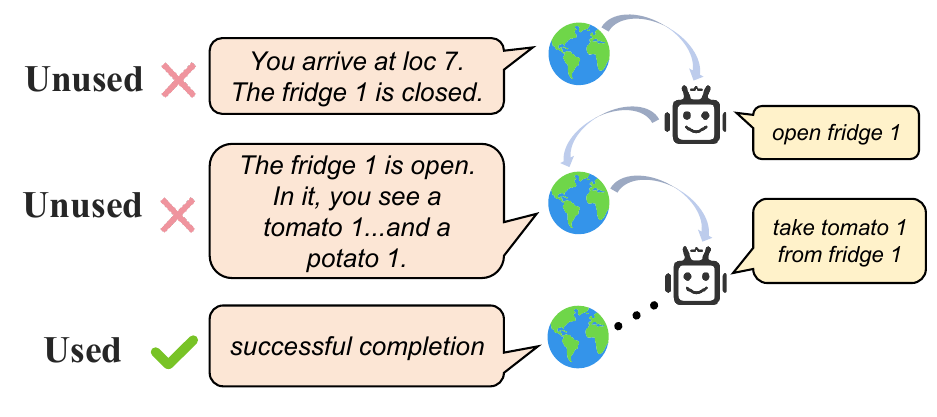}
    \caption{\textbf{Interaction process in agentic RL.} Existing agentic RL methods assign credit primarily from sparse, trajectory-level outcome rewards (e.g., “successful completion”), while the rich environmental feedback during interactions is under-explored.}
    \vspace{-6.0pt}
    \label{fig:figure1}
\end{wrapfigure}
Large language models (LLMs) are increasingly deployed as autonomous agents to handle complex tasks, ranging from utilizing software tools~\citep{singh2025openaigpt5card, 5team2025glm45agenticreasoningcoding} to interacting with the physical world~\citep{kim24openvla, pi0.5}. These tasks necessitate multi-turn interactions with the environment, demanding not only strong language understanding but also robust long-horizon planning capabilities derived from the model~\citep{gaia, zhou2023webarena, WebShop, ALFWorld}. Consequently, Reinforcement Learning (RL) has emerged as a standard paradigm to train LLMs as agents~\citep{jin2025searchr1, RAGEN, qin2025uitars}. By learning through trial-and-error, RL allows agents to explore the environment and discover effective strategies directly from environmental feedback, rather than relying solely on expert trajectories.

However, prevailing agentic RL algorithms, such as GRPO~\citep{DeepSeekMath} and RLOO~\citep{RLOO}, predominantly rely on outcome-based rewards, where the environment provides binary feedback only upon the completion of an episode. Such sparse environment feedback poses a severe challenge for learning in long-horizon, multi-turn contexts~\citep{agentRLsurvey, agenticRLsurvey2,RAGEN}. In contrast, we argue that the feedback available from the environment is much richer than mere outcomes. As shown in Figure~\ref{fig:figure1}, while the agent interacts with the environment, it generates experience that implicitly carries dense signals, describing how the environment evolves in response to actions and revealing the underlying transition mechanism~\citep{wm, LeCun2022APT}. Learning from these signals enables the agent to understand the dynamics of the environment and identify meaningful trajectories within vast state spaces~\citep{dreamer, plan2explore}. Motivated by this, we aim to investigate how to harness the environment dynamics contained in experience to enhance agentic RL.

In this work, we propose a simple yet effective framework, \textbf{\ourmethod{}}, that enables agents to continuously learn environment dynamics as extra supervision signals during RL updates. Concretely, we explore two strategies to transform experiences into self-supervision signals to introduce the environment dynamics for agentic RL. 1) \textbf{State-Prediction} (SP), which models forward environment dynamics by predicting the next state given the current state and action, encouraging the agent to capture how its actions shape subsequent observations. 2) \textbf{Inverse-Dynamics} (ID), which trains the agent to infer the action that caused a particular state transition, thereby strengthening the agent’s understanding of the causal link between its behavior and environmental changes. By jointly optimizing SP and ID alongside the RL objective, \ourmethod{} allows the agent to internalize environment dynamics from accumulated experience and make superior decisions. Importantly, \ourmethod{} reuses rollout data generated during RL and requires only a single additional forward pass per update, introducing negligible computational overhead.

We evaluate our method on two representative long-horizon agentic benchmarks, ALFWorld~\citep{ALFWorld} and WebShop~\citep{WebShop}, across multiple model scales and reinforcement learning baselines. Results show that \ourmethod{} consistently improves success rate over standard policy optimization methods, including GRPO~\citep{DeepSeekMath}, for both 1.5B and 7B models. On Qwen2.5‑1.5B-Instruct with GRPO, it yields +4.6\% on ALFWorld and +10.2\% on WebShop. Comparable gains hold at 7B and when applied on stronger baselines such as GiGPO~\citep{gigpo}, indicating complementary benefits beyond improved policy optimization alone. Overall, \ourmethod{} offers a lightweight and general approach for LLM agents to learn from environment dynamics, consistently boosting agentic RL across different settings.

\section{Preliminaries}
\label{sec:prelim}
\subsection{Task Formulation}
\label{sec:task_formulation}

We consider an agent interacting with a textual environment over discrete time steps, which can be formalized as a Partially Observable Markov Decision Process (POMDP) defined by $(\mathcal{S}, \mathcal{A}, \mathcal{P}, \mathcal{R}, \gamma)$. At each time step $t$, the agent receives a natural language observation $s_t \in \mathcal{S}$ and selects a textual action $a_t \in \mathcal{A}$ according to a policy $\pi_\theta(a_t \mid h_t)$, where $h_t = (s_0, a_0, \dots, s_t)$ denotes the interaction history up to time $t$ and $\theta$ denotes the trainable parameters of the underlying language model that parameterizes the policy. The environment then transitions to a new observation $s_{t+1}$ governed by the dynamics $\mathcal{P}(s_{t+1} \mid s_t, a_t)$. An episode corresponds to a multi-turn interaction trajectory $\tau = (s_0, a_0, s_1, \dots, s_T)$, representing the complete sequence of observations and actions exchanged between the agent and the environment, where $T$ denotes the episode horizon. An episode ends either upon successful task completion or when the maximum number of interaction steps $T$ is reached.

In agentic environments, an observation $s_t$ typically consists of textual descriptions of the current environment state, such as scene descriptions, available objects, or webpage information, while an action $a_t$ is expressed as a text-based command, typically selected from an environment-provided set of admissible actions that define the valid operations at the current state.

\subsection{Reinforcement Learning Objective}

We consider outcome-based reinforcement learning, where each trajectory $\tau$ is assigned a scalar reward $R(\tau)$ reflecting overall task success. We adopt Group Relative Policy Optimization (GRPO) as our baseline due to its critic-free design and stable advantage estimation~\citep{DeepSeekMath}. 

Specifically, the policy samples a group of $K$ trajectories $\{\tau^{(k)}\}_{k=1}^K$ with corresponding returns $\{R^{(k)}\}_{k=1}^K$. The RL objective is defined and formulated as:
\begin{equation}
\begin{aligned}
\mathcal{L}_{RL}(\theta) &= -\mathbb{E}_{k,t} \left[ \min \left( \frac{\pi_\theta(a_{k,t}\mid h_{k,t})}{\pi_{\theta_{\mathrm{old}}}(a_{k,t}\mid h_{k,t})}\tilde{A}_k, \operatorname{clip}\left(\frac{\pi_\theta(a_{k,t}\mid h_{k,t})}{\pi_{\theta_{\mathrm{old}}}(a_{k,t}\mid h_{k,t})}, 1-\epsilon, 1+\epsilon\right)\tilde{A}_k \right) \right] \\
&\quad + \beta \operatorname{KL}(\pi_\theta \| \pi_{\mathrm{ref}})
\end{aligned}
\end{equation}

where $\pi_\theta$ and $\pi_{\theta_{\mathrm{old}}}$ denote the current and the old policies, and $h_{k,t}$ represents the interaction history up to step $t$. $\tilde{A}_k$ is the efficiently computed group-relative advantage, calculated as $\tilde{A}_k = \frac{R_k-\mathrm{mean}(\{R_j\}_{j=1}^K)}{\mathrm{std}(\{R_j\}_{j=1}^K)+\epsilon}.$

    

Since our method is orthogonal to the choice of reinforcement learning algorithm, it can be combined with any policy optimization method. To verify the robustness of our approach against a stronger RL baseline, we additionally employ GiGPO~\citep{gigpo}, an advanced group-based policy optimization method featuring more fine-grained credit assignment. Results in Section~\ref{sec:exp} demonstrate that our method consistently yields further improvements even when built upon this stronger RL baseline.
\section{Method}

\begin{figure*}[t]
    \centering
    \includegraphics[width=1\linewidth]{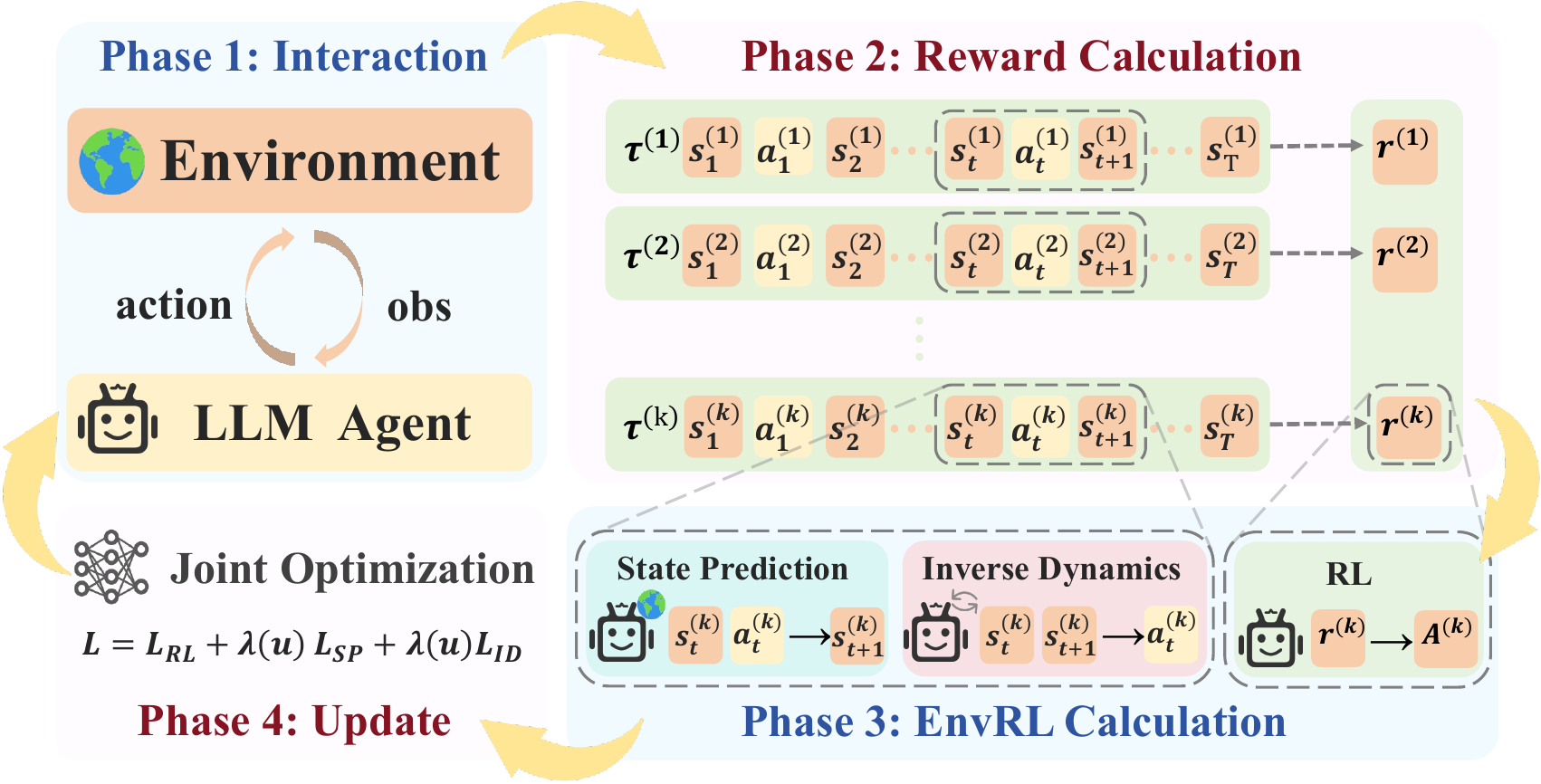}
    \caption{Overview of our \ourmethod{} framework. The LLM agent interacts with the environment to collect multiple rollouts, each producing an outcome reward. We then reuse the rollout trajectories to construct two auxiliary self-supervised training signals: \textbf{State Prediction~(SP)} trains the agent to predict how the environment state changes after an action, and \textbf{Inverse Dynamics~(ID)} trains the agent to infer which action caused a given state transition. In parallel, the outcome rewards are used for standard RL updates. Finally, we jointly optimize the agent with the RL objective and the two auxiliary objectives (with weights decaying over training), and repeat this process iteratively.}
    \label{fig:method}
\end{figure*}

\label{sec:method}

\subsection{Overview}
In this section, we present the design details of our method. Our central idea is to uncover the environment dynamics information as implicit supervision signals from the agent’s interaction experience and integrate them into the reinforcement learning (RL) process. Specifically, we design two objectives to achieve this. The first one is the \textbf{state-prediction} objective, which trains the agent to predict how the environment state will change, and the other one is the \textbf{inverse-dynamics} objective, which asks the agent to infer what action caused a given state transition. Finally, we treat these two objectives as auxiliary objectives and jointly optimize them together with the RL objective during training. The overall training process is shown in Figure~\ref{fig:method}.
\subsection{State Prediction}

 Learning predictive models of environment dynamics has long been a powerful paradigm in sequential decision-making. Prior work has shown that training agents to anticipate state transitions $p(s_{t+1}|s_t, a_t)$ enables them to simulate future trajectories and foresee the long-term consequences of their behaviors, deriving more robust policies~\citep{dreamer, muzero, dreamerv3}.

Motivated by this mechanism, we first  introduce the \textbf{State Prediction (SP)} objective. Given the current state $s_t$ and the chosen action $a_t$, the agent is trained to predict the next environment state $s_{t+1}$. In our setting, this state is instantiated as the next-step textual observation. By explicitly anticipating the consequences of its actions, the agent can internalize environment dynamics through its own interaction experience. Specifically, we formulate the SP learning objective as:
\begin{equation}
\mathcal{L}_{\mathrm{SP}}(\theta)
=
- \mathbb{E}_{(s_t,a_t,s_{t+1}) \sim \mathcal{D}_{\pi_\theta}}
\left[
\log p_\theta(s_{t+1} \mid s_t, a_t)
\right].
\label{eq:l_sp}
\end{equation}
Here, the transition tuples $(s_t, a_t, s_{t+1})$ are derived from rollout trajectories collected under the current policy $\pi_\theta$. Each trajectory $\tau = (s_0, a_0, s_1, \ldots, s_T)$ is decomposed into consecutive tuples $\{(s_t, a_t, s_{t+1})\}_{t=0}^{T-1}$, serving as self-supervised training samples. Our approach is fully on-policy: training data is generated by the current policy and discarded after each update. In practice, since $s_{t+1}$ is a textual observation, this objective reduces to the standard token-level cross-entropy loss, akin to supervised fine-tuning on self-generated interaction experience.

\subsection{Inverse Dynamics}

Although the SP objective provides valuable forward-looking supervision to help the agent understand how the environment evolves, this alone is insufficient. For effective decision-making, the agent must also reason backwards to understand exactly how its actions influence the environment. Prior work has shown that inverse dynamics---predicting the causal action between consecutive states---encourages representations to filter out extraneous observation details and focus strictly on agent-controllable aspects~\citep{curiosity, id}. Motivated by this, we introduce \textbf{Inverse Dynamics~(ID)} as a complementary objective. Given two consecutive states $(s_t, s_{t+1})$, the agent is trained to recover the action $a_t$ that caused the transition. We formulate the ID learning objective as: 
\begin{equation}
\mathcal{L}_{\mathrm{ID}}(\theta)
=
- \mathbb{E}_{(s_t,a_t,s_{t+1}) \sim \mathcal{D}_{\pi_\theta}}
\left[
\log p_\theta(a_t \mid s_t, s_{t+1})
\right].
\label{eq:l_id}
\end{equation}

Here, the transition tuples $(s_t, a_t, s_{t+1})$ are obtained in the same manner as those used in the SP objective. Together, SP and ID provide a bidirectional model of environment dynamics:
SP captures how actions shape future states, while ID captures which actions explain observed state changes.

\subsection{Joint Online Optimization}

We design a joint online optimization approach for the agent to dynamically update its understanding of the environment during the RL process. Specifically, we treat the SP and ID tasks as auxiliary objectives and combine them with the original RL objective to form the overall optimization objective. At each update step, the total learning objective is:
\begin{equation}
\mathcal{L}(\theta)
=
\mathcal{L}_{\mathrm{RL}}(\theta)
+ \lambda_{\mathrm{SP}}\, \mathcal{L}_{\mathrm{SP}}(\theta)
+ \lambda_{\mathrm{ID}}\, \mathcal{L}_{\mathrm{ID}}(\theta),
\label{eq:l_total}
\end{equation}
where $\lambda_{\mathrm{SP}}$ and $\lambda_{\mathrm{ID}}$ are hyperparameters controling the contribution of each auxiliary objective.

However, maintaining fixed auxiliary coefficients throughout training presents a dilemma: large weights can dominate the learning objective and hinder policy improvement in later stages when the agent should focus on reward maximization; conversely, small weights provide insufficient guidance during early training when environment dynamics are still being learned. To resolve this tension, we introduce a coefficient decay mechanism that gradually reduces the contribution of auxiliary objectives as training progresses, formulated as:
\begin{equation}
\mathcal{L}_{\mathrm{\ourmethod{}}}(\theta; t)
=
\mathcal{L}_{\mathrm{RL}}(\theta)
+
\lambda_{\mathrm{SP}}(t)\,\mathcal{L}_{\mathrm{SP}}(\theta)
+
\lambda_{\mathrm{ID}}(t)\,\mathcal{L}_{\mathrm{ID}}(\theta),
\end{equation}
where $\lambda_{\mathrm{SP}}(t)$ and $\lambda_{\mathrm{ID}}(t)$
are time-dependent coefficients, controlled by a cosine decay schedule:
\begin{equation}
\begin{aligned}
\lambda_{\mathrm{SP}}(t)
&=
\lambda_{\mathrm{SP}} \cdot
\frac{1}{2}\left(1+\cos\left(\pi \frac{t}{T_{max}}\right)\right),\\
\lambda_{\mathrm{ID}}(t)
&=
\lambda_{\mathrm{ID}} \cdot
\frac{1}{2}\left(1+\cos\left(\pi \frac{t}{T_{max}}\right)\right),
\end{aligned}
\end{equation}
where $t$ denotes the current training step and $T_{max}$ is the total number of training steps. This schedule ensures a smooth transition without abrupt changes in optimization dynamics.

With this design, the agent receives dense supervision from auxiliary objectives in the early phase, encouraging it to capture environment dynamics. As training proceeds, the auxiliary coefficients gradually decay, allowing the optimization to
smoothly shift toward the primary RL objective.

\section{Experiments}

\label{sec:exp}

\subsection{Experimental Setup}
\label{sec:exp_setup}

\paragraph{Benchmarks.} We evaluate our method on two widely-used long-horizon agentic benchmarks: ALFWorld and WebShop. ALFWorld~\citep{ALFWorld} is a text-based interactive environment that simulates household tasks. Each task requires the agent to complete multi-step object manipulation goals. The benchmark consists of six task types: \textit{Pick}, \textit{Look}, \textit{Clean}, \textit{Heat}, \textit{Cool}, and \textit{Pick2}. We use the standard train/test split and report success rates on the test set. Success is determined by whether the agent achieves the specified goal within the maximum allowed steps. WebShop \citep{WebShop} is a simulated e-commerce environment where agents must search and purchase products that match given user instructions. The task involves navigating a product catalog, and selecting the item that best matches the requirements. For ALFWorld, we compute the success rate for each task type, as well as the overall success rate across all task types. For WebShop, we report the score metric (measuring attribute-level matching between the purchased item and the instruction) and the overall success rate.

\paragraph{Baselines.} 
We compare against the following baselines. For prompting-based baselines, we include Prompting, which directly prompts the model with task instructions and represents zero-shot performance, and ReAct~\citep{react}, which interleaves reasoning traces (thoughts) and actions to generate explicit reasoning steps before taking actions. For RL-based baselines, we use GRPO~\citep{DeepSeekMath}, a group-based RL algorithm, as our primary baseline, and GiGPO \citep{gigpo}, a more advanced group-based policy optimization method with finer-grained credit assignment. Our method, denoted as \ourmethod{}, improves these RL baselines by incorporating the auxiliary objectives described in~\ref{sec:method}, noted as \ourmethod{}-GRPO and \ourmethod{}-GiGPO. For reference, we also include results from two powerful proprietary models, GPT-4o \citep{openai2024gpt4ocard} and Gemini-2.5-Pro \citep{comanici2025gemini25}, both evaluated using direct prompting. 

\paragraph{Implementation Details.}
We use Qwen2.5-1.5B/7B-Instruct~\citep{qwen2.5} as our base policy models. We follow the training configuration of the verl-agent framework~\citep{gigpo}. Specifically, we use a batch size of 16, set the group size to K=8 for group-based RL, and train for 150 steps with a learning rate of 1e-6 using AdamW optimizer. For our \ourmethod{} method, we additionally introduce auxiliary objectives with initial coefficients $\lambda_{SP}=\lambda_{ID}=0.2$ for 1.5B models and $\lambda_{SP}=\lambda_{ID}=0.1$ for 7B models. All experiments are conducted on 8$\times$A100 gpus. The prompts used are listed in Appendix~\ref{appendix:prompts}.

\begin{table*}[t]
\centering
\small
\caption{Main results of \ourmethod{}. We report task-wise success rates (\%) for six sub-tasks and the overall average (All) for ALFWorld, and performance in terms of Score and Success rate (Succ.) for WebShop. Results are averaged over 3 random seeds, with standard deviations shown where possible.}

\label{tab:main}
\setlength\tabcolsep{4pt}
\renewcommand{\arraystretch}{1}
\resizebox{\textwidth}{!}{%
\begin{tabular}{p{2.8cm}l ccccccc cc}
\toprule
\multirow{2}[2]{*}{\textbf{Method}} & \multicolumn{7}{c}{\textbf{ALFWorld}} & \multicolumn{2}{c}{\textbf{WebShop}} \\
\cmidrule(lr){2-8} \cmidrule(lr){9-10}

& Pick & Look & Clean & Heat & Cool & Pick2 & All
& Score & Succ. \\
\midrule
\multicolumn{10}{c}{\textit{\textbf{Backbone Model: Proprietary Model}}} \\
\midrule
GPT-4o        &  75.3 & 60.8 & 31.2 & 56.7 & 21.6 & 49.8 & 48.0 & 31.8 & 23.7 \\
Gemini-2.5-Pro &  92.8 & 63.3 & 62.1 & 69.0 & 26.6 & 58.7 & 60.3 & 42.5 & 35.9 \\

\midrule
\multicolumn{10}{c}{\textit{\textbf{Backbone Model: Qwen2.5-1.5B-Instruct}}} \\
\midrule

  Prompting
  & 5.9 & 5.5 & 3.3 & 9.7 & 4.2 & 0.0 & 4.1 & 23.1 & 5.2 \\
  ReAct & 17.4 & 20.5 & 15.7 & 6.2 & 7.7 & 2.0 & 12.8 & 40.1 & 11.3 \\
  GRPO
  & 85.3$_{\pm1.5}$ & 53.7$_{\pm8.0}$ & 84.5$_{\pm6.8}$
  & 78.2$_{\pm7.9}$ & 59.7$_{\pm5.0}$ & 53.5$_{\pm5.6}$
  & 72.8$_{\pm3.6}$
  & 75.8$_{\pm3.5}$ & 56.8$_{\pm3.8}$ \\
  \rowcolor[HTML]{f0edff}\ourmethod-GRPO
  & \textbf{85.9}$_{\pm2.3}$ & \textbf{53.9}$_{\pm6.9}$ & \textbf{85.0}$_{\pm4.5}$ 
  & \textbf{82.0}$_{\pm4.7}$ & \textbf{76.8}$_{\pm3.9}$ & \textbf{71.9}$_{\pm4.4}$ 
  & \textbf{77.4}$_{\pm3.8}$ 
  & \textbf{83.0}$_{\pm2.1}$ & \textbf{67.0}$_{\pm2.6}$ \\
  GiGPO
  & 94.4$_{\pm5.9}$ & 67.5$_{\pm4.6}$ & 94.8$_{\pm3.8}$
  & 94.4$_{\pm7.8}$ & 79.8$_{\pm4.7}$ & 76.4$_{\pm5.4}$
  & 86.7$_{\pm1.7}$
  & 83.5$_{\pm1.8}$ & 67.4$_{\pm4.5}$ \\
 \rowcolor[HTML]{f0edff} \ourmethod-GiGPO
  & \textbf{95.0}$_{\pm3.3}$ & \textbf{70.8}$_{\pm6.4}$ & \textbf{94.9}$_{\pm4.6}$ 
  & \textbf{97.7}$_{\pm3.5}$ & \textbf{94.1}$_{\pm4.8}$ & \textbf{94.7}$_{\pm4.7}$ 
  & \textbf{91.8}$_{\pm3.5}$ 
  & \textbf{88.2}$_{\pm2.5}$ & \textbf{74.2}$_{\pm3.7}$ \\

\midrule
\multicolumn{10}{c}{\textit{\textbf{Backbone Model: Qwen2.5-7B-Instruct}}} \\
\midrule
  Prompting
  & 33.4 & 21.6 & 19.3 & 6.9 & 2.8 & 3.2 & 14.8 & 26.4 & 7.8 \\
  ReAct & 48.5 & 35.4 & 34.3 & 13.2 & 18.2 & 17.6 & 31.2 & 46.2 & 19.5 \\
  GRPO
  & 90.8$_{\pm5.1}$ & 66.1$_{\pm6.7}$ & 89.3$_{\pm5.4}$
  & 74.7$_{\pm6.9}$ & 72.5$_{\pm5.4}$ & 64.7$_{\pm7.3}$
  & 77.6$_{\pm5.2}$
  & 79.3$_{\pm2.8}$ & 66.1$_{\pm3.7}$ \\
  \rowcolor[HTML]{f0edff} \ourmethod-GRPO
  & \textbf{92.6}$_{\pm4.3}$ & \textbf{77.5}$_{\pm5.4}$ & \textbf{92.9}$_{\pm3.9}$ 
  & \textbf{79.7}$_{\pm3.0}$ & \textbf{77.4}$_{\pm3.7}$ & \textbf{71.1}$_{\pm6.2}$ 
  & \textbf{80.4}$_{\pm3.9}$ 
  & \textbf{81.8}$_{\pm1.9}$ & \textbf{68.6}$_{\pm2.3}$ \\
  GiGPO
  & 97.7$_{\pm1.6}$ & 82.7$_{\pm7.9}$ & 98.8$_{\pm1.6}$
  & 83.7$_{\pm7.2}$ & 89.3$_{\pm8.2}$ & 79.2$_{\pm6.6}$
  & 90.8$_{\pm1.3}$
  & 86.2$_{\pm2.6}$ & 75.2$_{\pm3.8}$ \\
  \rowcolor[HTML]{f0edff} \ourmethod-GiGPO
  & \textbf{98.3}$_{\pm1.5}$ & \textbf{92.2}$_{\pm4.5}$ & \textbf{98.9}$_{\pm0.8}$ 
  & \textbf{93.8}$_{\pm6.9}$ & \textbf{93.3}$_{\pm3.5}$ & \textbf{94.6}$_{\pm3.0}$ 
  & \textbf{94.5}$_{\pm4.0}$ 
  & \textbf{88.4}$_{\pm4.9}$ & \textbf{76.3}$_{\pm4.7}$ \\
\bottomrule
\vspace{1em}
\end{tabular}
}
\end{table*}

\subsection{Main Results}

Table~\ref{tab:main} demonstrates the effectiveness of incorporating the auxiliary objectives into RL training for long-horizon agentic tasks. While powerful proprietary models (e.g., GPT-4o, Gemini-2.5-Pro) and prompting-based approaches struggle on benchmarks like ALFWorld and WebShop, outcome-based RL methods like GRPO and GiGPO substantially improve agent capabilities. For instance, on the 1.5B model, GiGPO achieves an 86.7\% success rate on ALFWorld and 67.4\% on WebShop. However, standard RL lacks environment dynamics modeling, limiting its ability to fully exploit supervision signals from the environment. \ourmethod{} addresses this limitation by jointly optimizing the SP and ID auxiliary objectives with RL, yielding consistent improvements across baseline algorithms and model scales. On the 1.5B model, \ourmethod{}-GRPO improves ALFWorld success rate from 72.8\% to 77.4\% (+4.6 points) and WebShop from 56.8\% to 67.0\% (+10.2 points). Improvement remains consistent with GiGPO, the stronger RL baseline. GiGPO + \ourmethod{} reaches 91.8\% on ALFWorld (+5.1 points) and 74.2\% on WebShop (+6.8 points). These benefits persist at the 7B scale, with GiGPO + \ourmethod{} achieving peak performances of 94.5\% and 76.3\%, respectively. Ultimately, these results validate that learning from environment dynamics provides valuable training signals that effectively complement sparse task rewards, enabling the agent to make better decisions in long-horizon tasks.

\subsection{Ablation Study}
\label{sec:ablation}

\begin{wraptable}{r}{0.45\textwidth}
\vspace{-1.3em}
\centering
\caption{Ablation study of \ourmethod{}. We conduct experiments by respectively removing SP, removing ID, removing the decay schedule, and applying a linear decay schedule~(instead of the cosine schedule).}
\label{tab:ablation}
\begin{adjustbox}{width=\linewidth}
\begin{tabular}{lcc}
\toprule
\textbf{Setting} & \textbf{ALFWorld} & \textbf{WebShop} \\
\midrule
\ourmethod{}-GRPO & \textbf{77.4} & \textbf{67.0} \\
\hspace{1em}w/o ID    & 73.5 & 64.5 \\
\hspace{1em}w/o SP    & 71.9 & 66.0  \\
\hspace{1em}w/o decay & 74.2 & 62.8  \\
\hspace{1em}w/ linear decay & 76.3 & 64.2  \\
\midrule
\ourmethod{}-GiGPO & \textbf{91.8} & \textbf{74.2} \\
\hspace{1em}w/o ID    & 89.3 & 71.4 \\
\hspace{1em}w/o SP    & 86.3 & 72.9  \\
\hspace{1em}w/o decay & 88.5 & 71.1  \\
\hspace{1em}w/ linear decay & 90.5 & 72.3  \\
\bottomrule
\end{tabular}
\end{adjustbox}
\vspace{-0.3em}
\end{wraptable}


\paragraph{Impact of auxiliary objectives}
We conduct an ablation study on both benchmarks and RL baselines. Starting from the full \ourmethod{}, we remove each component individually while keeping all other settings fixed. We choose Qwen2.5-1.5B-Instruct as the base policy model. Results are shown in Table~\ref{tab:ablation}. On ALFWorld, removing SP causes the largest drop ($77.4\!\rightarrow\!71.9$ for GRPO and $91.8\!\rightarrow\!86.3$ for GiGPO). we attribute this to the highly logical state transitions in this environment, which requires rigorous reasoning to make precise predictions. In contrast, removing ID has a relatively minor impact because ALFWorld observations implicitly encode action information, thereby reducing the difficulty of inverse reasoning.
On WebShop, where observations are noisy web pages, ID is more influential than SP (e.g., $67.0\!\rightarrow\!64.5$ and $74.2\!\rightarrow\!71.4$ when removing ID), suggesting the importance for agents to identify meaningful actions from complex and informative observations. 

\paragraph{Impact of decay mechanism}

\begin{figure*}[ht]
    \centering
    \includegraphics[width=\textwidth]{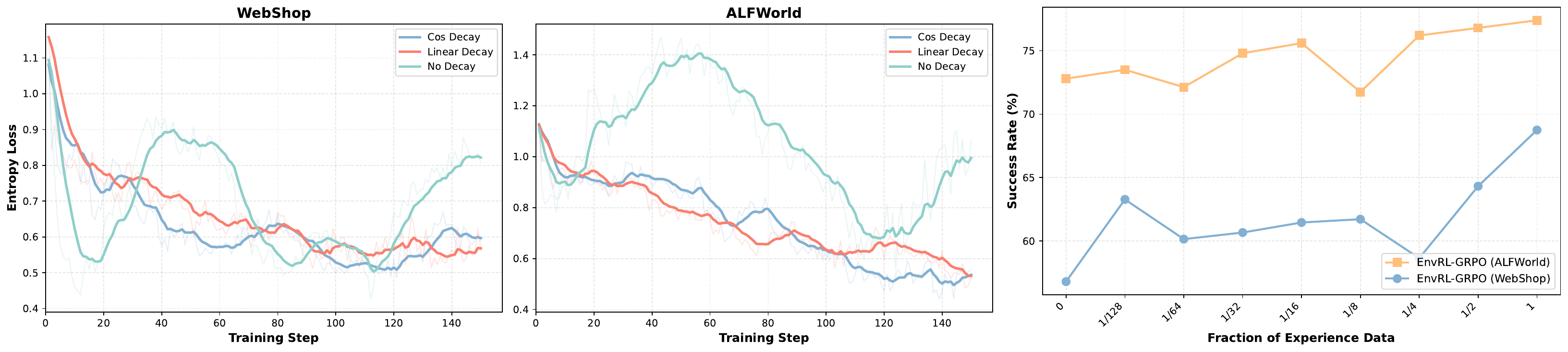}
    \caption{Empirical analysis of \ourmethod{}. \textit{Left:} Effect of decay schedules (cosine, linear, and no decay) on entropy loss during RL training. \textit{Right:} Effect of the fraction of experience data used for training on agent success rate across ALFWorld and WebShop     
  environments. }
    \label{fig:ablation_figure}
\end{figure*}

As shown in Table~\ref{tab:ablation}, disabling the decay mechanism consistently degrades performance (with a drop of up to 4.2 points on WebShop), indicating that perpetually optimizing auxiliary tasks ultimately hinders the learning of the primary objective. This is further corroborated by the entropy loss dynamics observed during RL training, as depicted in Figure~\ref{fig:ablation_figure}~(left). Without decay, the entropy loss remains persistently unstable, preventing the policy from fully converging on the main task. In contrast, applying a decay mechanism allows the entropy to smoothly converge to a stable minimum, ensuring the agent properly fine-tunes its policy. Furthermore, we evaluated a linear decay schedule, which performs slightly worse than cosine decay but remains superior to the no-decay baseline. We attribute this gap to the rapid early-stage reduction of the auxiliary weight under linear decay, leading to insufficient learning of the environment dynamics during the initial phase. Consequently, we adopt the cosine decay schedule as default for our method.

\paragraph{Impact of Experience Data Scale}

\ourmethod{} enables the agent to learn from environment dynamics by reformulating experience data into SP and ID tasks. To investigate how the scale of data affects the learning , we evaluate our model using varying proportions of the total experience data to form the SP and ID objective. Specifically, we use 1/128, 1/64, 1/32, 1/16, 1/8, 1/4, and 1/2 of the full experience dataset to train \ourmethod{}-GRPO, comparing them against baselines trained with RL only (fraction = 0) and with full experience data (fraction = 1). All experiments are conducted on the Qwen2.5-1.5B-Instruct model. Results are visualized in Figure~\ref{fig:ablation_figure}~(right). The results largely align with our expectations: performance generally improves as more experience data is utilized. On both benchmarks, we observe a consistent upward trend in success rate as the data fraction increases. Notably, even a small fraction of experience data (e.g., 1/128) already provides measurable gains over the RL-only baseline, and performance continues to improve as more data becomes available. This suggests that the auxiliary objectives effectively extract useful supervision from experience, and the benefits scale with the amount of experience data. 

\subsection{Analysis}

\begin{figure*}[t]
    \centering
    \includegraphics[width=\textwidth]{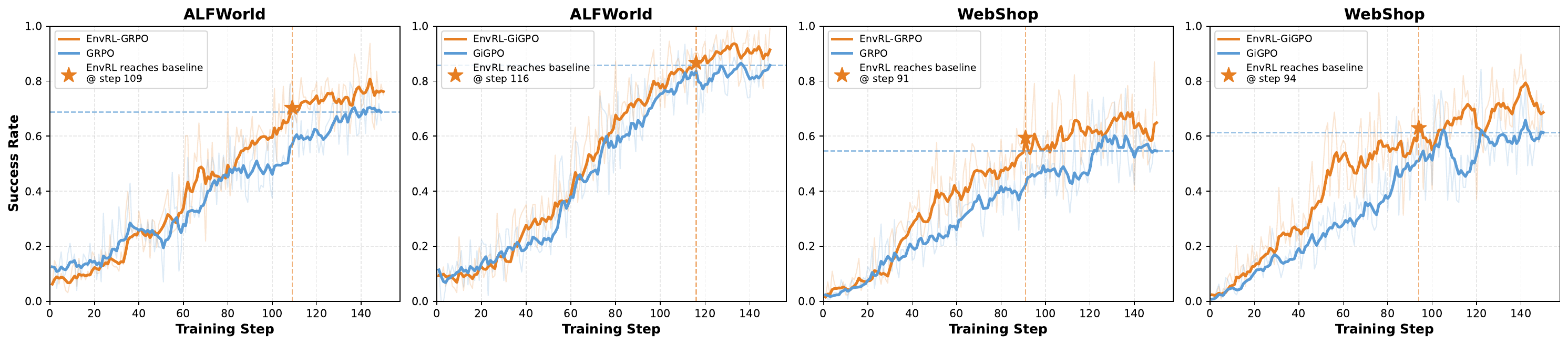}
    \caption{Success rate over training iterations. The orange lines represent our method, and the blue lines represents the original RL baseline.}
    \label{fig:success_rate}
\end{figure*}

\paragraph{Sample Efficiency.}                                                                                                
To evaluate the sample efficiency of \ourmethod{}, we report the success rate during the RL training process in Figure~\ref{fig:success_rate}. Augmenting both GRPO and GiGPO with \ourmethod{} consistently yields faster performance  
gains and higher final performance on both ALFWorld and WebShop. To quantify sample efficiency, we identify the earliest training step at which \ourmethod{} first reaches the final performance     
level of the RL baseline (marked with $\bigstar$ in the figure). Across all four settings, \ourmethod{} achieves equivalent performance to the fully-trained RL baseline using only $\sim$68.5\% of the total training steps on average (ranging from 60.7\% to 77.9\%), demonstrating that \ourmethod{} not only improves the        
performance ceiling but also substantially accelerates convergence. Interestingly, sometimes in the early stage of training, \ourmethod{} slightly lags behind the RL baseline. We attribute th is to the relatively large coefficients of the auxiliary objectives at the beginning, during which the model prioritizes learning environment dynamics. As the coefficients gradually decay, the model quickly shifts back to optimizing task rewards and catches up, which aligns with our expectations.

\begin{figure*}[t]
    \centering
    \includegraphics[width=\textwidth]{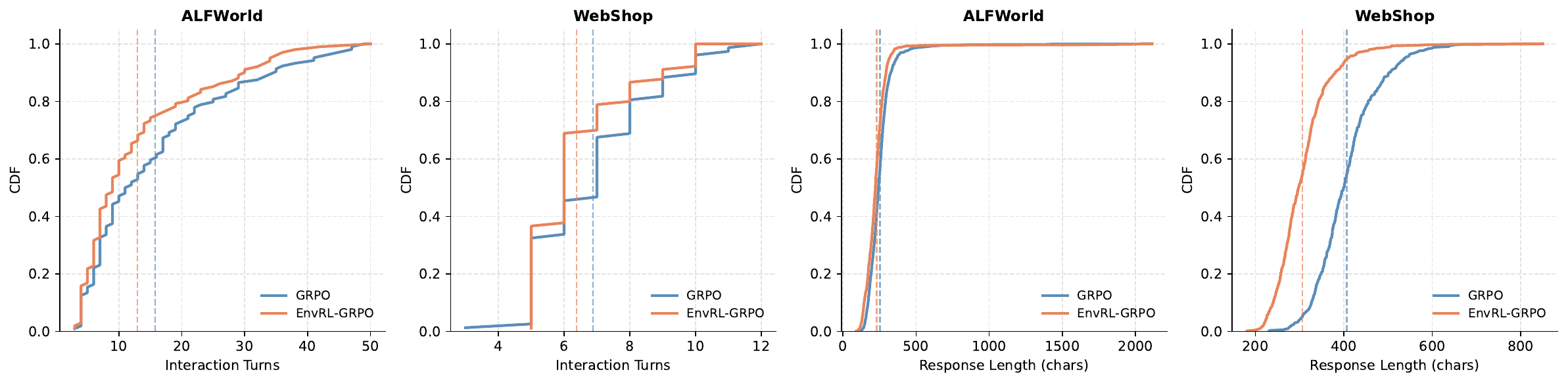}
    \caption{Cumulative Distribution Function (CDF) of interaction turns (left) and response lengths (right, measured in characters) for successful episodes on ALFWorld and WebShop. Dashed vertical lines indicate the mean values of the respective distributions. Across both environments, \ourmethod{}-GRPO consistently shifts the distributions to the left compared to the GRPO baseline, demonstrating that it achieves task success with fewer interactions and more concise textual outputs.}
    \label{fig:traj_cdf}
\end{figure*}

\paragraph{Inference Efficiency.} 
To further understand the behavioral improvements brought by our method, we plot the Cumulative Distribution Function (CDF) of interaction turns and response lengths for successful episodes in Figure~\ref{fig:traj_cdf}. The leftward shift of curves across all sub-figures clearly indicates that our method requires fewer interaction turns and shorter responses to solve tasks. The reduced number of turns shows that learning from environment dynamics deepens the agent's understanding of action-state causalities, rendering its decision-making more precise and cutting down on redundant exploratory behaviors. Similarly, shorter response lengths highlight that the agent becomes more confident and decisive, producing effective actions without excessive thinking overhead.

\begin{table}[t]
    \centering
    \caption{Out-of-Distribution (OOD) evaluation on ALFWorld, measured by success rate (\%).}
    \label{tab:alfworld_ood}
    \begin{tabular}{lcccc}
        \toprule
        \multirow{2}{*}{\textbf{Backbone Model}} & \multicolumn{2}{c}{\textbf{GRPO}} & \multicolumn{2}{c}{\textbf{GiGPO}} \\
        \cmidrule(lr){2-3} \cmidrule(lr){4-5}
        & Baseline & +\ourmethod{} & Baseline & +\ourmethod{} \\
        \midrule
        Qwen2.5-1.5B-Instruct & 61.19 & \textbf{64.43} & 75.12 & \textbf{78.84} \\
        Qwen2.5-7B-Instruct   & 66.45 & \textbf{70.18} & 79.23 & \textbf{81.76} \\
        \bottomrule
    \end{tabular}
\end{table}

\paragraph{Hyperparameter Sensitivity.}
\begin{wraptable}{r}{0.49\textwidth}
  \vspace{-1em}
    \centering
    \caption{Hyperparameter sensitivity analysis of the initial auxiliary coefficient $\lambda$.}
    \label{tab:hyperparam_sensitivity}
    \begin{adjustbox}{width=\linewidth}
    \begin{tabular}{llcc}
        \toprule
        \textbf{Model Size} & $\lambda$ & \textbf{ALFWorld} & \textbf{WebShop} \\
        \midrule
        \multirow{4}{*}{1.5B} & $\lambda=0$   & 72.8 & 56.8 \\
                       & $\lambda=0.1$ & 76.3 & 64.9 \\
                       & $\lambda=0.2$ & \textbf{77.4} & \textbf{67.0} \\
                       & $\lambda=0.4$ & 72.3 & 57.8 \\
        \midrule
        \multirow{4}{*}{7B}  & $\lambda=0$   & 77.6 & 66.1 \\
                       & $\lambda=0.1$ & \textbf{81.9} & \textbf{68.6} \\
                       & $\lambda=0.2$ & 80.0 & 67.3 \\
                       & $\lambda=0.4$ & 78.7 & 66.2 \\
        \bottomrule
    \end{tabular}
    \end{adjustbox}
\end{wraptable}
Despite the decay mechanism of the auxiliary coefficient, the initial value is critical for the whole optimization process. We conduct a grid search for the two initial coefficients, $\lambda_{SP}$ and $\lambda_{ID}$, to justify our choices and analyze its sensitivity. Based on our initial findings that individual tuning of these two hyperparameters yields marginal differences, for simplicity, we maintain $\lambda_{SP} = \lambda_{ID} = \lambda$ throughout our experiments. As shown in Table~\ref{tab:hyperparam_sensitivity}, we can conclude that an excessively high $\lambda$ hinders the learning of the primary task, whereas a $\lambda$ that is too low leads to insufficient learning of environment dynamics. The 1.5B model peaks at $\lambda = 0.2$, while the 7B model peaks at $\lambda = 0.1$. We attribute this to the stronger environmental priors in larger models, thus requiring less auxiliary supervision. Overall, it shows that the performance is relatively robust to the choice of this hyperparameter within the $[0.1, 0.2]$ range.

\paragraph{Generalization and Task Versatility.} 
To verify that \ourmethod{} captures underlying environment dynamics rather than overfitting to specific tasks, we evaluate its performance on ALFWorld OOD (Out-of-Distribution) scenarios.  Specifically, we employ the models trained on the standard ALFWorld training set and evaluate them directly on the OOD test set, which features entirely novel room configurations and unseen object types~\citep{ALFWorld}. As shown in Table~\ref{tab:alfworld_ood}, \ourmethod{} maintains a significant margin over the baseline, confirming that internalizing environment dynamics provides robust priors that generalize to unseen configurations. Furthermore, to extend the diversity of tasks, we conduct additional experiments on the SearchQA benchmark, which focuses on information-seeking scenarios. Detailed numerical results and setup for SearchQA experiments are provided in Appendix~\ref{appendix:searchqa}.

\vspace{1em}

\paragraph{Computational Efficiency}

\begin{wraptable}{r}{0.5\textwidth}
    \vspace{-1.0pt}
    \centering
    \includegraphics[width=0.5\textwidth]{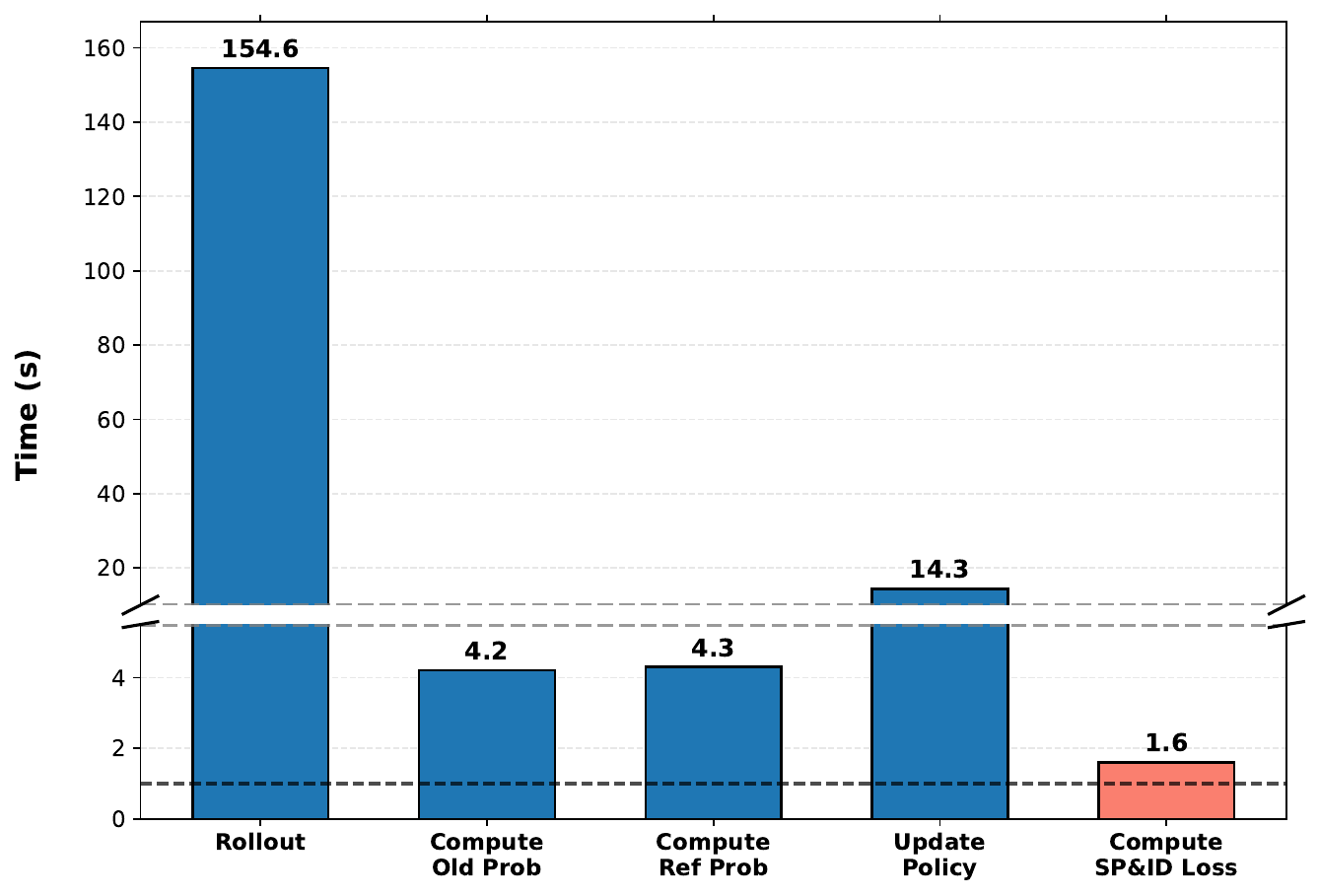}
    \caption{Time cost of RL training components. We report the average time spent on environment rollouts, probability computations, policy updates, and our auxiliary objective calculations.}
    \vspace{-5.0pt}
    \label{fig:time_cost}
\end{wraptable}

To quantify the computational efficiency of our method, we measure the time breakdown of each training component and report their average time cost during training, as shown in Figure~\ref{fig:time_cost}. The overall training time is dominated by environment interaction (\textsc{Rollout}, 154.6s) and policy optimization (\textsc{Update Policy}, 14.3s). In contrast, our auxiliary objectives introduce only a negligible overhead (1.6s). This is because \ourmethod{} reuses the rollout trajectories already generated by RL and requires only a single additional forward pass per update to compute the state-prediction and inverse-dynamics losses, without additional sampling or separate model training. As a result, \ourmethod{} incurs negligible extra computational overhead in practice.

\section{Related Works}

\paragraph{Agentic Reinforcement Learning.} 
Reinforcement Learning (RL) has evolved from a basic alignment tool for single-turn tasks into a core framework for enabling LLMs to act as autonomous agents in complex environments. Early research applied classical algorithms like DQN\citep{DQN} and PPO~\citep{PPO} to discrete environments such as games~\citep{alphago} and control~\citep{lillicrap2019continuouscontroldeepreinforcement}. To tackle complex real-world tasks such as web navigation~\citep{WebShop,zhou2023webarena}, software operation~\citep{OSWorld}, and multi-tool orchestration~\citep{wei2025browsecompsimplechallengingbenchmark, gaia}, group-based RL methods such as GRPO~\citep{DeepSeekMath} are being increasingly adapted~\citep{jin2025searchr1, RAGEN} for their computational efficiency and stable advantage estimation. However, a fundamental challenge in these multi-turn, long-horizon scenarios is the sparsity of environmental feedback, which provides limited supervision signals and leads to insufficient agent training~\citep{sparse}. To address this, existing research has has made significant strides in identifying intermediate states to develop fine-grained credit assignment schemes~\citep{gigpo, li2025saltstepleveladvantageassignment, VAGEN, liu2025agenticreinforcementlearningimplicit}. While effective, our work is largely orthogonal to these methods, focusing on mining implicit supervision signals directly from environment dynamics. This perspective treats interaction experience as a rich, independent information source, allowing our method to complement reward-centric RL methods and achieve superior performance through their integration.

\paragraph{Learning from Environment Dynamics}
Effective decision-making in long-horizon tasks requires agents to move beyond relying solely on sparse outcome rewards~\citep{agentsurvey1, agenticRLsurvey2}. By learning from environment dynamics, agents can exploit the rich structural information latent in their interaction trajectories, which reveals the underlying transition mechanisms of the environment\citep{wm, LeCun2022APT}. World models offer a practical way to achieve this by serving as explicit internal simulators that enable agents to anticipate state transitions and plan actions through latent imagination, thereby significantly enhancing sample efficiency and generalization~\citep{dreamer, dreamerv3, muzero, EarlyExp}. Our work is closely related to this line of research. We aim to enable agents to capture transition mechanisms through exploration in new environments, and treat it as additional supervision signal to facilitate superior decision-making.



\section{Conclusions}

We present \ourmethod{}, a RL framework that enables LLM agents to learn from environment dynamics for more efficient decision-making. \ourmethod{} integrates two auxiliary objectives---state prediction and inverse dynamics---with primary RL objective, and jointly optimizes them with a weight-decay mechanism. Empirically, \ourmethod{} consistently improves performance across different long-horizon agentic environments, model scales, and RL baselines, demonstrating its generalizability and complementarity to agentic RL algorithms. We hope this work can provide a new perspective on agentic RL mechanics, contributing to develop more robust, sample-efficient, and autonomous agents capable of increasingly complex real-world interactions.

\bibliographystyle{acl_natbib}
\bibliography{anthology,custom}

@article{EarlyExp,
  title        = {Agent Learning via Early Experience},
  author       = {Zhang, Kai and Chen, Xiangchao and Liu, Bo and others},
  year         = {2025},
  eprint       = {2510.08558},
  archivePrefix= {arXiv},
  primaryClass = {cs.AI},
  url          = {https://arxiv.org/abs/2510.08558}
}

@article{VAGEN,
  title        = {VAGEN: Reinforcing World Model Reasoning for Multi-Turn VLM Agents},
  author       = {Wang, Kangrui and Zhang, Pingyue and Wang, Zihan and others},
  year         = {2025},
  eprint       = {2510.16907},
  archivePrefix= {arXiv},
  primaryClass = {cs.AI},
  url          = {https://arxiv.org/abs/2510.16907}
}

@article{WebShop,
  title        = {WebShop: Towards Scalable Real-World Web Interaction with Grounded Language Agents},
  author       = {Yao, Shunyu and Chen, Howard and Yang, John and Narasimhan, Karthik},
  year         = {2022},
  eprint       = {2207.01206},
  archivePrefix= {arXiv},
  primaryClass = {cs.CL},
  doi          = {10.48550/arXiv.2207.01206},
  url          = {https://arxiv.org/abs/2207.01206}
}

@article{ALFWorld,
  title        = {ALFWorld: Aligning Text and Embodied Environments for Interactive Learning},
  author       = {Shridhar, Mohit and Yuan, Xingdi and C{\^o}t{\'e}, Marc-Alexandre and Bisk, Yonatan and Trischler, Adam and Hausknecht, Matthew},
  year         = {2020},
  eprint       = {2010.03768},
  archivePrefix= {arXiv},
  primaryClass = {cs.CL},
  url          = {https://arxiv.org/abs/2010.03768}
}

@article{DeepSeekMath,
  title        = {DeepSeekMath: Pushing the Limits of Mathematical Reasoning in Open Language Models},
  author       = {Shao, Zhihong and Wang, Peiyi and Zhu, Qihao and Xu, Runxin and Song, Junxiao and Bi, Xiao and Zhang, Haowei and Zhang, Mingchuan and Li, Y. K. and Wu, Y. and Guo, Daya},
  year         = {2024},
  eprint       = {2402.03300},
  archivePrefix= {arXiv},
  primaryClass = {cs.CL},
  doi          = {10.48550/arXiv.2402.03300},
  url          = {https://arxiv.org/abs/2402.03300}
}

@misc{react,
Author = {Shunyu Yao and Jeffrey Zhao and Dian Yu and Nan Du and Izhak Shafran and Karthik Narasimhan and Yuan Cao},
Title = {ReAct: Synergizing Reasoning and Acting in Language Models},
Year = {2022},
Eprint = {arXiv:2210.03629},
}

@misc{dreamer,
Author = {Danijar Hafner and Timothy Lillicrap and Jimmy Ba and Mohammad Norouzi},
Title = {Dream to Control: Learning Behaviors by Latent Imagination},
Year = {2019},
Eprint = {arXiv:1912.01603},
}

@article{wm,
Author = {David Ha and Jürgen Schmidhuber},
Title = {World Models},
Year = {2018},
Eprint = {arXiv:1803.10122},
Doi = {10.5281/zenodo.1207631},
}

@misc{agentRLsurvey,
Author = {Guibin Zhang and Hejia Geng and Xiaohang Yu and Zhenfei Yin and Zaibin Zhang and Zelin Tan and Heng Zhou and Zhongzhi Li and Xiangyuan Xue and Yijiang Li and Yifan Zhou and Yang Chen and Chen Zhang and Yutao Fan and Zihu Wang and Songtao Huang and Francisco Piedrahita-Velez and Yue Liao and Hongru Wang and Mengyue Yang and Heng Ji and Jun Wang and Shuicheng Yan and Philip Torr and Lei Bai},
Title = {The Landscape of Agentic Reinforcement Learning for LLMs: A Survey},
Year = {2025},
Eprint = {arXiv:2509.02547},
}

@misc{RAGEN,
Author = {Zihan Wang and Kangrui Wang and Qineng Wang and Pingyue Zhang and Linjie Li and Zhengyuan Yang and Xing Jin and Kefan Yu and Minh Nhat Nguyen and Licheng Liu and Eli Gottlieb and Yiping Lu and Kyunghyun Cho and Jiajun Wu and Li Fei-Fei and Lijuan Wang and Yejin Choi and Manling Li},
Title = {RAGEN: Understanding Self-Evolution in LLM Agents via Multi-Turn Reinforcement Learning},
Year = {2025},
Eprint = {arXiv:2504.20073},
}

@article{zhou2023webarena,
  title={WebArena: A Realistic Web Environment for Building Autonomous Agents},
  author={Zhou, Shuyan and Xu, Frank F and Zhu, Hao and Zhou, Xuhui and Lo, Robert and Sridhar, Abishek and Cheng, Xianyi and Bisk, Yonatan and Fried, Daniel and Alon, Uri and others},
  journal={arXiv preprint arXiv:2307.13854},
  url={https://webarena.dev},
  year={2023}
}

@article{kim24openvla,
    title={OpenVLA: An Open-Source Vision-Language-Action Model},
    author={{Moo Jin} Kim and Karl Pertsch and Siddharth Karamcheti and Ted Xiao and Ashwin Balakrishna and Suraj Nair and Rafael Rafailov and Ethan Foster and Grace Lam and Pannag Sanketi and Quan Vuong and Thomas Kollar and Benjamin Burchfiel and Russ Tedrake and Dorsa Sadigh and Sergey Levine and Percy Liang and Chelsea Finn},
    journal = {arXiv preprint arXiv:2406.09246},
    year={2024}
}

@misc{pi0.5,
Author = {Physical Intelligence and Kevin Black and Noah Brown and James Darpinian and Karan Dhabalia and Danny Driess and Adnan Esmail and Michael Equi and Chelsea Finn and Niccolo Fusai and Manuel Y. Galliker and Dibya Ghosh and Lachy Groom and Karol Hausman and Brian Ichter and Szymon Jakubczak and Tim Jones and Liyiming Ke and Devin LeBlanc and Sergey Levine and Adrian Li-Bell and Mohith Mothukuri and Suraj Nair and Karl Pertsch and Allen Z. Ren and Lucy Xiaoyang Shi and Laura Smith and Jost Tobias Springenberg and Kyle Stachowicz and James Tanner and Quan Vuong and Homer Walke and Anna Walling and Haohuan Wang and Lili Yu and Ury Zhilinsky},
Title = {$\pi_{0.5}$: a Vision-Language-Action Model with Open-World Generalization},
Year = {2025},
Eprint = {arXiv:2504.16054},
}

@article{gigpo,
  title={Group-in-Group Policy Optimization for LLM Agent Training},
  author={Feng, Lang and Xue, Zhenghai and Liu, Tingcong and An, Bo},
  journal={arXiv preprint arXiv:2505.10978},
  year={2025}
}

@misc{gaia,
Author = {Grégoire Mialon and Clémentine Fourrier and Craig Swift and Thomas Wolf and Yann LeCun and Thomas Scialom},
Title = {GAIA: a benchmark for General AI Assistants},
Year = {2023},
Eprint = {arXiv:2311.12983},
}

@misc{5team2025glm45agenticreasoningcoding,
      title={GLM-4.5: Agentic, Reasoning, and Coding (ARC) Foundation Models}, 
      author={GLM Team and Aohan Zeng and Xin Lv and Qinkai Zheng and Zhenyu Hou and Bin Chen and Chengxing Xie and Cunxiang Wang and Da Yin and Hao Zeng and Jiajie Zhang and Kedong Wang and Lucen Zhong and Mingdao Liu and Rui Lu and Shulin Cao and Xiaohan Zhang and Xuancheng Huang and Yao Wei and Yean Cheng and Yifan An and Yilin Niu and Yuanhao Wen and Yushi Bai and Zhengxiao Du and Zihan Wang and Zilin Zhu and Bohan Zhang and Bosi Wen and Bowen Wu and Bowen Xu and Can Huang and Casey Zhao and Changpeng Cai and Chao Yu and Chen Li and Chendi Ge and Chenghua Huang and Chenhui Zhang and Chenxi Xu and Chenzheng Zhu and Chuang Li and Congfeng Yin and Daoyan Lin and Dayong Yang and Dazhi Jiang and Ding Ai and Erle Zhu and Fei Wang and Gengzheng Pan and Guo Wang and Hailong Sun and Haitao Li and Haiyang Li and Haiyi Hu and Hanyu Zhang and Hao Peng and Hao Tai and Haoke Zhang and Haoran Wang and Haoyu Yang and He Liu and He Zhao and Hongwei Liu and Hongxi Yan and Huan Liu and Huilong Chen and Ji Li and Jiajing Zhao and Jiamin Ren and Jian Jiao and Jiani Zhao and Jianyang Yan and Jiaqi Wang and Jiayi Gui and Jiayue Zhao and Jie Liu and Jijie Li and Jing Li and Jing Lu and Jingsen Wang and Jingwei Yuan and Jingxuan Li and Jingzhao Du and Jinhua Du and Jinxin Liu and Junkai Zhi and Junli Gao and Ke Wang and Lekang Yang and Liang Xu and Lin Fan and Lindong Wu and Lintao Ding and Lu Wang and Man Zhang and Minghao Li and Minghuan Xu and Mingming Zhao and Mingshu Zhai and Pengfan Du and Qian Dong and Shangde Lei and Shangqing Tu and Shangtong Yang and Shaoyou Lu and Shijie Li and Shuang Li and Shuang-Li and Shuxun Yang and Sibo Yi and Tianshu Yu and Wei Tian and Weihan Wang and Wenbo Yu and Weng Lam Tam and Wenjie Liang and Wentao Liu and Xiao Wang and Xiaohan Jia and Xiaotao Gu and Xiaoying Ling and Xin Wang and Xing Fan and Xingru Pan and Xinyuan Zhang and Xinze Zhang and Xiuqing Fu and Xunkai Zhang and Yabo Xu and Yandong Wu and Yida Lu and Yidong Wang and Yilin Zhou and Yiming Pan and Ying Zhang and Yingli Wang and Yingru Li and Yinpei Su and Yipeng Geng and Yitong Zhu and Yongkun Yang and Yuhang Li and Yuhao Wu and Yujiang Li and Yunan Liu and Yunqing Wang and Yuntao Li and Yuxuan Zhang and Zezhen Liu and Zhen Yang and Zhengda Zhou and Zhongpei Qiao and Zhuoer Feng and Zhuorui Liu and Zichen Zhang and Zihan Wang and Zijun Yao and Zikang Wang and Ziqiang Liu and Ziwei Chai and Zixuan Li and Zuodong Zhao and Wenguang Chen and Jidong Zhai and Bin Xu and Minlie Huang and Hongning Wang and Juanzi Li and Yuxiao Dong and Jie Tang},
      year={2025},
      eprint={2508.06471},
      archivePrefix={arXiv},
      primaryClass={cs.CL},
      url={https://arxiv.org/abs/2508.06471}, 
}

@misc{singh2025openaigpt5card,
      title={OpenAI GPT-5 System Card}, 
      author={Aaditya Singh and Adam Fry and Adam Perelman and Adam Tart and Adi Ganesh and Ahmed El-Kishky and Aidan McLaughlin and Aiden Low and AJ Ostrow and Akhila Ananthram and Akshay Nathan and Alan Luo and Alec Helyar and Aleksander Madry and Aleksandr Efremov and Aleksandra Spyra and Alex Baker-Whitcomb and Alex Beutel and Alex Karpenko and Alex Makelov and Alex Neitz and Alex Wei and Alexandra Barr and Alexandre Kirchmeyer and Alexey Ivanov and Alexi Christakis and Alistair Gillespie and Allison Tam and Ally Bennett and Alvin Wan and Alyssa Huang and Amy McDonald Sandjideh and Amy Yang and Ananya Kumar and Andre Saraiva and Andrea Vallone and Andrei Gheorghe and Andres Garcia Garcia and Andrew Braunstein and Andrew Liu and Andrew Schmidt and Andrey Mereskin and Andrey Mishchenko and Andy Applebaum and Andy Rogerson and Ann Rajan and Annie Wei and Anoop Kotha and Anubha Srivastava and Anushree Agrawal and Arun Vijayvergiya and Ashley Tyra and Ashvin Nair and Avi Nayak and Ben Eggers and Bessie Ji and Beth Hoover and Bill Chen and Blair Chen and Boaz Barak and Borys Minaiev and Botao Hao and Bowen Baker and Brad Lightcap and Brandon McKinzie and Brandon Wang and Brendan Quinn and Brian Fioca and Brian Hsu and Brian Yang and Brian Yu and Brian Zhang and Brittany Brenner and Callie Riggins Zetino and Cameron Raymond and Camillo Lugaresi and Carolina Paz and Cary Hudson and Cedric Whitney and Chak Li and Charles Chen and Charlotte Cole and Chelsea Voss and Chen Ding and Chen Shen and Chengdu Huang and Chris Colby and Chris Hallacy and Chris Koch and Chris Lu and Christina Kaplan and Christina Kim and CJ Minott-Henriques and Cliff Frey and Cody Yu and Coley Czarnecki and Colin Reid and Colin Wei and Cory Decareaux and Cristina Scheau and Cyril Zhang and Cyrus Forbes and Da Tang and Dakota Goldberg and Dan Roberts and Dana Palmie and Daniel Kappler and Daniel Levine and Daniel Wright and Dave Leo and David Lin and David Robinson and Declan Grabb and Derek Chen and Derek Lim and Derek Salama and Dibya Bhattacharjee and Dimitris Tsipras and Dinghua Li and Dingli Yu and DJ Strouse and Drew Williams and Dylan Hunn and Ed Bayes and Edwin Arbus and Ekin Akyurek and Elaine Ya Le and Elana Widmann and Eli Yani and Elizabeth Proehl and Enis Sert and Enoch Cheung and Eri Schwartz and Eric Han and Eric Jiang and Eric Mitchell and Eric Sigler and Eric Wallace and Erik Ritter and Erin Kavanaugh and Evan Mays and Evgenii Nikishin and Fangyuan Li and Felipe Petroski Such and Filipe de Avila Belbute Peres and Filippo Raso and Florent Bekerman and Foivos Tsimpourlas and Fotis Chantzis and Francis Song and Francis Zhang and Gaby Raila and Garrett McGrath and Gary Briggs and Gary Yang and Giambattista Parascandolo and Gildas Chabot and Grace Kim and Grace Zhao and Gregory Valiant and Guillaume Leclerc and Hadi Salman and Hanson Wang and Hao Sheng and Haoming Jiang and Haoyu Wang and Haozhun Jin and Harshit Sikchi and Heather Schmidt and Henry Aspegren and Honglin Chen and Huida Qiu and Hunter Lightman and Ian Covert and Ian Kivlichan and Ian Silber and Ian Sohl and Ibrahim Hammoud and Ignasi Clavera and Ikai Lan and Ilge Akkaya and Ilya Kostrikov and Irina Kofman and Isak Etinger and Ishaan Singal and Jackie Hehir and Jacob Huh and Jacqueline Pan and Jake Wilczynski and Jakub Pachocki and James Lee and James Quinn and Jamie Kiros and Janvi Kalra and Jasmyn Samaroo and Jason Wang and Jason Wolfe and Jay Chen and Jay Wang and Jean Harb and Jeffrey Han and Jeffrey Wang and Jennifer Zhao and Jeremy Chen and Jerene Yang and Jerry Tworek and Jesse Chand and Jessica Landon and Jessica Liang and Ji Lin and Jiancheng Liu and Jianfeng Wang and Jie Tang and Jihan Yin and Joanne Jang and Joel Morris and Joey Flynn and Johannes Ferstad and Johannes Heidecke and John Fishbein and John Hallman and Jonah Grant and Jonathan Chien and Jonathan Gordon and Jongsoo Park and Jordan Liss and Jos Kraaijeveld and Joseph Guay and Joseph Mo and Josh Lawson and Josh McGrath and Joshua Vendrow and Joy Jiao and Julian Lee and Julie Steele and Julie Wang and Junhua Mao and Kai Chen and Kai Hayashi and Kai Xiao and Kamyar Salahi and Kan Wu and Karan Sekhri and Karan Sharma and Karan Singhal and Karen Li and Kenny Nguyen and Keren Gu-Lemberg and Kevin King and Kevin Liu and Kevin Stone and Kevin Yu and Kristen Ying and Kristian Georgiev and Kristie Lim and Kushal Tirumala and Kyle Miller and Lama Ahmad and Larry Lv and Laura Clare and Laurance Fauconnet and Lauren Itow and Lauren Yang and Laurentia Romaniuk and Leah Anise and Lee Byron and Leher Pathak and Leon Maksin and Leyan Lo and Leyton Ho and Li Jing and Liang Wu and Liang Xiong and Lien Mamitsuka and Lin Yang and Lindsay McCallum and Lindsey Held and Liz Bourgeois and Logan Engstrom and Lorenz Kuhn and Louis Feuvrier and Lu Zhang and Lucas Switzer and Lukas Kondraciuk and Lukasz Kaiser and Manas Joglekar and Mandeep Singh and Mandip Shah and Manuka Stratta and Marcus Williams and Mark Chen and Mark Sun and Marselus Cayton and Martin Li and Marvin Zhang and Marwan Aljubeh and Matt Nichols and Matthew Haines and Max Schwarzer and Mayank Gupta and Meghan Shah and Melody Huang and Meng Dong and Mengqing Wang and Mia Glaese and Micah Carroll and Michael Lampe and Michael Malek and Michael Sharman and Michael Zhang and Michele Wang and Michelle Pokrass and Mihai Florian and Mikhail Pavlov and Miles Wang and Ming Chen and Mingxuan Wang and Minnia Feng and Mo Bavarian and Molly Lin and Moose Abdool and Mostafa Rohaninejad and Nacho Soto and Natalie Staudacher and Natan LaFontaine and Nathan Marwell and Nelson Liu and Nick Preston and Nick Turley and Nicklas Ansman and Nicole Blades and Nikil Pancha and Nikita Mikhaylin and Niko Felix and Nikunj Handa and Nishant Rai and Nitish Keskar and Noam Brown and Ofir Nachum and Oleg Boiko and Oleg Murk and Olivia Watkins and Oona Gleeson and Pamela Mishkin and Patryk Lesiewicz and Paul Baltescu and Pavel Belov and Peter Zhokhov and Philip Pronin and Phillip Guo and Phoebe Thacker and Qi Liu and Qiming Yuan and Qinghua Liu and Rachel Dias and Rachel Puckett and Rahul Arora and Ravi Teja Mullapudi and Raz Gaon and Reah Miyara and Rennie Song and Rishabh Aggarwal and RJ Marsan and Robel Yemiru and Robert Xiong and Rohan Kshirsagar and Rohan Nuttall and Roman Tsiupa and Ronen Eldan and Rose Wang and Roshan James and Roy Ziv and Rui Shu and Ruslan Nigmatullin and Saachi Jain and Saam Talaie and Sam Altman and Sam Arnesen and Sam Toizer and Sam Toyer and Samuel Miserendino and Sandhini Agarwal and Sarah Yoo and Savannah Heon and Scott Ethersmith and Sean Grove and Sean Taylor and Sebastien Bubeck and Sever Banesiu and Shaokyi Amdo and Shengjia Zhao and Sherwin Wu and Shibani Santurkar and Shiyu Zhao and Shraman Ray Chaudhuri and Shreyas Krishnaswamy and Shuaiqi and Xia and Shuyang Cheng and Shyamal Anadkat and Simón Posada Fishman and Simon Tobin and Siyuan Fu and Somay Jain and Song Mei and Sonya Egoian and Spencer Kim and Spug Golden and SQ Mah and Steph Lin and Stephen Imm and Steve Sharpe and Steve Yadlowsky and Sulman Choudhry and Sungwon Eum and Suvansh Sanjeev and Tabarak Khan and Tal Stramer and Tao Wang and Tao Xin and Tarun Gogineni and Taya Christianson and Ted Sanders and Tejal Patwardhan and Thomas Degry and Thomas Shadwell and Tianfu Fu and Tianshi Gao and Timur Garipov and Tina Sriskandarajah and Toki Sherbakov and Tomer Kaftan and Tomo Hiratsuka and Tongzhou Wang and Tony Song and Tony Zhao and Troy Peterson and Val Kharitonov and Victoria Chernova and Vineet Kosaraju and Vishal Kuo and Vitchyr Pong and Vivek Verma and Vlad Petrov and Wanning Jiang and Weixing Zhang and Wenda Zhou and Wenlei Xie and Wenting Zhan and Wes McCabe and Will DePue and Will Ellsworth and Wulfie Bain and Wyatt Thompson and Xiangning Chen and Xiangyu Qi and Xin Xiang and Xinwei Shi and Yann Dubois and Yaodong Yu and Yara Khakbaz and Yifan Wu and Yilei Qian and Yin Tat Lee and Yinbo Chen and Yizhen Zhang and Yizhong Xiong and Yonglong Tian and Young Cha and Yu Bai and Yu Yang and Yuan Yuan and Yuanzhi Li and Yufeng Zhang and Yuguang Yang and Yujia Jin and Yun Jiang and Yunyun Wang and Yushi Wang and Yutian Liu and Zach Stubenvoll and Zehao Dou and Zheng Wu and Zhigang Wang},
      year={2025},
      eprint={2601.03267},
      archivePrefix={arXiv},
      primaryClass={cs.CL},
      url={https://arxiv.org/abs/2601.03267}, 
}

@article{jin2025searchr1,
  title={Search-r1: Training llms to reason and leverage search engines with reinforcement learning},
  author={Jin, Bowen and Zeng, Hansi and Yue, Zhenrui and Yoon, Jinsung and Arik, Sercan and Wang, Dong and Zamani, Hamed and Han, Jiawei},
  journal={arXiv preprint arXiv:2503.09516},
  year={2025}
}

@article{qin2025uitars,
  title={UI-TARS: Pioneering Automated GUI Interaction with Native Agents},
  author={Qin, Yujia and Ye, Yining and Fang, Junjie and Wang, Haoming and Liang, Shihao and Tian, Shizuo and Zhang, Junda and Li, Jiahao and Li, Yunxin and Huang, Shijue and others},
  journal={arXiv preprint arXiv:2501.12326},
  year={2025}
}

@misc{agenticRLsurvey2,
Author = {Minhua Lin and Zongyu Wu and Zhichao Xu and Hui Liu and Xianfeng Tang and Qi He and Charu Aggarwal and Hui Liu and Xiang Zhang and Suhang Wang},
Title = {A Comprehensive Survey on Reinforcement Learning-based Agentic Search: Foundations, Roles, Optimizations, Evaluations, and Applications},
Year = {2025},
Eprint = {arXiv:2510.16724},
}

@misc{RLOO,
Author = {Arash Ahmadian and Chris Cremer and Matthias Gallé and Marzieh Fadaee and Julia Kreutzer and Olivier Pietquin and Ahmet Üstün and Sara Hooker},
Title = {Back to Basics: Revisiting REINFORCE Style Optimization for Learning from Human Feedback in LLMs},
Year = {2024},
Eprint = {arXiv:2402.14740},
}

@inproceedings{LeCun2022APT,
  title={A Path Towards Autonomous Machine Intelligence Version 0.9.2, 2022-06-27},
  author={Yann LeCun and Courant},
  year={2022},
  url={https://api.semanticscholar.org/CorpusID:251881108}
}

@misc{plan2explore,
Author = {Ramanan Sekar and Oleh Rybkin and Kostas Daniilidis and Pieter Abbeel and Danijar Hafner and Deepak Pathak},
Title = {Planning to Explore via Self-Supervised World Models},
Year = {2020},
Eprint = {arXiv:2005.05960},
}

@article{muzero,
Author = {Julian Schrittwieser and Ioannis Antonoglou and Thomas Hubert and Karen Simonyan and Laurent Sifre and Simon Schmitt and Arthur Guez and Edward Lockhart and Demis Hassabis and Thore Graepel and Timothy Lillicrap and David Silver},
Title = {Mastering Atari, Go, Chess and Shogi by Planning with a Learned Model},
Year = {2019},
Eprint = {arXiv:1911.08265},
Doi = {10.1038/s41586-020-03051-4},
}

@InProceedings{curiosity,
  title = 	 {Curiosity-driven Exploration by Self-supervised Prediction},
  author =       {Deepak Pathak and Pulkit Agrawal and Alexei A. Efros and Trevor Darrell},
  booktitle = 	 {Proceedings of the 34th International Conference on Machine Learning},
  pages = 	 {2778--2787},
  year = 	 {2017},
  editor = 	 {Precup, Doina and Teh, Yee Whye},
  volume = 	 {70},
  series = 	 {Proceedings of Machine Learning Research},
  month = 	 {06--11 Aug},
  publisher =    {PMLR},
  url = 	 {https://proceedings.mlr.press/v70/pathak17a.html}
}

@inproceedings{id,
author = {Brandfonbrener, David and Nachum, Ofir and Bruna, Joan},
title = {Inverse dynamics pretraining learns good representations for multitask imitation},
year = {2023},
publisher = {Curran Associates Inc.},
address = {Red Hook, NY, USA},
booktitle = {Proceedings of the 37th International Conference on Neural Information Processing Systems},
articleno = {2925},
numpages = {26},
location = {New Orleans, LA, USA},
series = {NIPS '23}
}

@article{qwen2.5,
    title   = {Qwen2.5 Technical Report}, 
    author  = {An Yang and Baosong Yang and Beichen Zhang and Binyuan Hui and Bo Zheng and Bowen Yu and Chengyuan Li and Dayiheng Liu and Fei Huang and Haoran Wei and Huan Lin and Jian Yang and Jianhong Tu and Jianwei Zhang and Jianxin Yang and Jiaxi Yang and Jingren Zhou and Junyang Lin and Kai Dang and Keming Lu and Keqin Bao and Kexin Yang and Le Yu and Mei Li and Mingfeng Xue and Pei Zhang and Qin Zhu and Rui Men and Runji Lin and Tianhao Li and Tingyu Xia and Xingzhang Ren and Xuancheng Ren and Yang Fan and Yang Su and Yichang Zhang and Yu Wan and Yuqiong Liu and Zeyu Cui and Zhenru Zhang and Zihan Qiu},
    journal = {arXiv preprint arXiv:2412.15115},
    year    = {2024}
}

@misc{openai2024gpt4ocard,
      title={GPT-4o System Card}, 
      author={OpenAI and : and Aaron Hurst and Adam Lerer and Adam P. Goucher and Adam Perelman and Aditya Ramesh and Aidan Clark and AJ Ostrow and Akila Welihinda and Alan Hayes and Alec Radford and Aleksander Mądry and Alex Baker-Whitcomb and Alex Beutel and Alex Borzunov and Alex Carney and Alex Chow and Alex Kirillov and Alex Nichol and Alex Paino and Alex Renzin and Alex Tachard Passos and Alexander Kirillov and Alexi Christakis and Alexis Conneau and Ali Kamali and Allan Jabri and Allison Moyer and Allison Tam and Amadou Crookes and Amin Tootoochian and Amin Tootoonchian and Ananya Kumar and Andrea Vallone and Andrej Karpathy and Andrew Braunstein and Andrew Cann and Andrew Codispoti and Andrew Galu and Andrew Kondrich and Andrew Tulloch and Andrey Mishchenko and Angela Baek and Angela Jiang and Antoine Pelisse and Antonia Woodford and Anuj Gosalia and Arka Dhar and Ashley Pantuliano and Avi Nayak and Avital Oliver and Barret Zoph and Behrooz Ghorbani and Ben Leimberger and Ben Rossen and Ben Sokolowsky and Ben Wang and Benjamin Zweig and Beth Hoover and Blake Samic and Bob McGrew and Bobby Spero and Bogo Giertler and Bowen Cheng and Brad Lightcap and Brandon Walkin and Brendan Quinn and Brian Guarraci and Brian Hsu and Bright Kellogg and Brydon Eastman and Camillo Lugaresi and Carroll Wainwright and Cary Bassin and Cary Hudson and Casey Chu and Chad Nelson and Chak Li and Chan Jun Shern and Channing Conger and Charlotte Barette and Chelsea Voss and Chen Ding and Cheng Lu and Chong Zhang and Chris Beaumont and Chris Hallacy and Chris Koch and Christian Gibson and Christina Kim and Christine Choi and Christine McLeavey and Christopher Hesse and Claudia Fischer and Clemens Winter and Coley Czarnecki and Colin Jarvis and Colin Wei and Constantin Koumouzelis and Dane Sherburn and Daniel Kappler and Daniel Levin and Daniel Levy and David Carr and David Farhi and David Mely and David Robinson and David Sasaki and Denny Jin and Dev Valladares and Dimitris Tsipras and Doug Li and Duc Phong Nguyen and Duncan Findlay and Edede Oiwoh and Edmund Wong and Ehsan Asdar and Elizabeth Proehl and Elizabeth Yang and Eric Antonow and Eric Kramer and Eric Peterson and Eric Sigler and Eric Wallace and Eugene Brevdo and Evan Mays and Farzad Khorasani and Felipe Petroski Such and Filippo Raso and Francis Zhang and Fred von Lohmann and Freddie Sulit and Gabriel Goh and Gene Oden and Geoff Salmon and Giulio Starace and Greg Brockman and Hadi Salman and Haiming Bao and Haitang Hu and Hannah Wong and Haoyu Wang and Heather Schmidt and Heather Whitney and Heewoo Jun and Hendrik Kirchner and Henrique Ponde de Oliveira Pinto and Hongyu Ren and Huiwen Chang and Hyung Won Chung and Ian Kivlichan and Ian O'Connell and Ian O'Connell and Ian Osband and Ian Silber and Ian Sohl and Ibrahim Okuyucu and Ikai Lan and Ilya Kostrikov and Ilya Sutskever and Ingmar Kanitscheider and Ishaan Gulrajani and Jacob Coxon and Jacob Menick and Jakub Pachocki and James Aung and James Betker and James Crooks and James Lennon and Jamie Kiros and Jan Leike and Jane Park and Jason Kwon and Jason Phang and Jason Teplitz and Jason Wei and Jason Wolfe and Jay Chen and Jeff Harris and Jenia Varavva and Jessica Gan Lee and Jessica Shieh and Ji Lin and Jiahui Yu and Jiayi Weng and Jie Tang and Jieqi Yu and Joanne Jang and Joaquin Quinonero Candela and Joe Beutler and Joe Landers and Joel Parish and Johannes Heidecke and John Schulman and Jonathan Lachman and Jonathan McKay and Jonathan Uesato and Jonathan Ward and Jong Wook Kim and Joost Huizinga and Jordan Sitkin and Jos Kraaijeveld and Josh Gross and Josh Kaplan and Josh Snyder and Joshua Achiam and Joy Jiao and Joyce Lee and Juntang Zhuang and Justyn Harriman and Kai Fricke and Kai Hayashi and Karan Singhal and Katy Shi and Kavin Karthik and Kayla Wood and Kendra Rimbach and Kenny Hsu and Kenny Nguyen and Keren Gu-Lemberg and Kevin Button and Kevin Liu and Kiel Howe and Krithika Muthukumar and Kyle Luther and Lama Ahmad and Larry Kai and Lauren Itow and Lauren Workman and Leher Pathak and Leo Chen and Li Jing and Lia Guy and Liam Fedus and Liang Zhou and Lien Mamitsuka and Lilian Weng and Lindsay McCallum and Lindsey Held and Long Ouyang and Louis Feuvrier and Lu Zhang and Lukas Kondraciuk and Lukasz Kaiser and Luke Hewitt and Luke Metz and Lyric Doshi and Mada Aflak and Maddie Simens and Madelaine Boyd and Madeleine Thompson and Marat Dukhan and Mark Chen and Mark Gray and Mark Hudnall and Marvin Zhang and Marwan Aljubeh and Mateusz Litwin and Matthew Zeng and Max Johnson and Maya Shetty and Mayank Gupta and Meghan Shah and Mehmet Yatbaz and Meng Jia Yang and Mengchao Zhong and Mia Glaese and Mianna Chen and Michael Janner and Michael Lampe and Michael Petrov and Michael Wu and Michele Wang and Michelle Fradin and Michelle Pokrass and Miguel Castro and Miguel Oom Temudo de Castro and Mikhail Pavlov and Miles Brundage and Miles Wang and Minal Khan and Mira Murati and Mo Bavarian and Molly Lin and Murat Yesildal and Nacho Soto and Natalia Gimelshein and Natalie Cone and Natalie Staudacher and Natalie Summers and Natan LaFontaine and Neil Chowdhury and Nick Ryder and Nick Stathas and Nick Turley and Nik Tezak and Niko Felix and Nithanth Kudige and Nitish Keskar and Noah Deutsch and Noel Bundick and Nora Puckett and Ofir Nachum and Ola Okelola and Oleg Boiko and Oleg Murk and Oliver Jaffe and Olivia Watkins and Olivier Godement and Owen Campbell-Moore and Patrick Chao and Paul McMillan and Pavel Belov and Peng Su and Peter Bak and Peter Bakkum and Peter Deng and Peter Dolan and Peter Hoeschele and Peter Welinder and Phil Tillet and Philip Pronin and Philippe Tillet and Prafulla Dhariwal and Qiming Yuan and Rachel Dias and Rachel Lim and Rahul Arora and Rajan Troll and Randall Lin and Rapha Gontijo Lopes and Raul Puri and Reah Miyara and Reimar Leike and Renaud Gaubert and Reza Zamani and Ricky Wang and Rob Donnelly and Rob Honsby and Rocky Smith and Rohan Sahai and Rohit Ramchandani and Romain Huet and Rory Carmichael and Rowan Zellers and Roy Chen and Ruby Chen and Ruslan Nigmatullin and Ryan Cheu and Saachi Jain and Sam Altman and Sam Schoenholz and Sam Toizer and Samuel Miserendino and Sandhini Agarwal and Sara Culver and Scott Ethersmith and Scott Gray and Sean Grove and Sean Metzger and Shamez Hermani and Shantanu Jain and Shengjia Zhao and Sherwin Wu and Shino Jomoto and Shirong Wu and Shuaiqi and Xia and Sonia Phene and Spencer Papay and Srinivas Narayanan and Steve Coffey and Steve Lee and Stewart Hall and Suchir Balaji and Tal Broda and Tal Stramer and Tao Xu and Tarun Gogineni and Taya Christianson and Ted Sanders and Tejal Patwardhan and Thomas Cunninghman and Thomas Degry and Thomas Dimson and Thomas Raoux and Thomas Shadwell and Tianhao Zheng and Todd Underwood and Todor Markov and Toki Sherbakov and Tom Rubin and Tom Stasi and Tomer Kaftan and Tristan Heywood and Troy Peterson and Tyce Walters and Tyna Eloundou and Valerie Qi and Veit Moeller and Vinnie Monaco and Vishal Kuo and Vlad Fomenko and Wayne Chang and Weiyi Zheng and Wenda Zhou and Wesam Manassra and Will Sheu and Wojciech Zaremba and Yash Patil and Yilei Qian and Yongjik Kim and Youlong Cheng and Yu Zhang and Yuchen He and Yuchen Zhang and Yujia Jin and Yunxing Dai and Yury Malkov},
      year={2024},
      eprint={2410.21276},
      archivePrefix={arXiv},
      primaryClass={cs.CL},
      url={https://arxiv.org/abs/2410.21276}, 
}

@misc{comanici2025gemini25,
  title         = {Gemini 2.5: Pushing the Frontier with Advanced Reasoning, Multimodality, Long Context, and Next Generation Agentic Capabilities},
  author        = {Gheorghe Comanici and Eric Bieber and Mike Schaekermann and Ice Pasupat and Noveen Sachdeva and Inderjit Dhillon and others},
  year          = {2025},
  eprint        = {2507.06261},
  archivePrefix = {arXiv},
  primaryClass  = {cs.CL},
  url           = {https://arxiv.org/abs/2507.06261}
}

@article{DQN,
  author       = {Volodymyr Mnih and
                  Koray Kavukcuoglu and
                  David Silver and
                  Alex Graves and
                  Ioannis Antonoglou and
                  Daan Wierstra and
                  Martin A. Riedmiller},
  title        = {Playing Atari with Deep Reinforcement Learning},
  journal      = {CoRR},
  volume       = {abs/1312.5602},
  year         = {2013},
  url          = {http://arxiv.org/abs/1312.5602},
  eprinttype    = {arXiv},
  eprint       = {1312.5602},
  bibsource    = {dblp computer science bibliography, https://dblp.org}
}

@article{PPO,
  author       = {John Schulman and
                  Filip Wolski and
                  Prafulla Dhariwal and
                  Alec Radford and
                  Oleg Klimov},
  title        = {Proximal Policy Optimization Algorithms},
  journal      = {CoRR},
  volume       = {abs/1707.06347},
  year         = {2017},
  url          = {http://arxiv.org/abs/1707.06347},
  eprinttype    = {arXiv},
  eprint       = {1707.06347},
  bibsource    = {dblp computer science bibliography, https://dblp.org}
}

@article{alphago,
  author       = {David Silver and
                  Thomas Hubert and
                  Julian Schrittwieser and
                  Ioannis Antonoglou and
                  Matthew Lai and
                  Arthur Guez and
                  Marc Lanctot and
                  Laurent Sifre and
                  Dharshan Kumaran and
                  Thore Graepel and
                  Timothy P. Lillicrap and
                  Karen Simonyan and
                  Demis Hassabis},
  title        = {Mastering Chess and Shogi by Self-Play with a General Reinforcement
                  Learning Algorithm},
  journal      = {CoRR},
  volume       = {abs/1712.01815},
  year         = {2017},
  url          = {http://arxiv.org/abs/1712.01815},
  eprinttype    = {arXiv},
  eprint       = {1712.01815},
  bibsource    = {dblp computer science bibliography, https://dblp.org}
}

@misc{lillicrap2019continuouscontroldeepreinforcement,
      title={Continuous control with deep reinforcement learning}, 
      author={Timothy P. Lillicrap and Jonathan J. Hunt and Alexander Pritzel and Nicolas Heess and Tom Erez and Yuval Tassa and David Silver and Daan Wierstra},
      year={2019},
      eprint={1509.02971},
      archivePrefix={arXiv},
      primaryClass={cs.LG},
      url={https://arxiv.org/abs/1509.02971}, 
}

@misc{wei2025browsecompsimplechallengingbenchmark,
      title={BrowseComp: A Simple Yet Challenging Benchmark for Browsing Agents}, 
      author={Jason Wei and Zhiqing Sun and Spencer Papay and Scott McKinney and Jeffrey Han and Isa Fulford and Hyung Won Chung and Alex Tachard Passos and William Fedus and Amelia Glaese},
      year={2025},
      eprint={2504.12516},
      archivePrefix={arXiv},
      primaryClass={cs.CL},
      url={https://arxiv.org/abs/2504.12516}, 
}

@misc{OSWorld,
      title={OSWorld: Benchmarking Multimodal Agents for Open-Ended Tasks in Real Computer Environments}, 
      author={Tianbao Xie and Danyang Zhang and Jixuan Chen and Xiaochuan Li and Siheng Zhao and Ruisheng Cao and Toh Jing Hua and Zhoujun Cheng and Dongchan Shin and Fangyu Lei and Yitao Liu and Yiheng Xu and Shuyan Zhou and Silvio Savarese and Caiming Xiong and Victor Zhong and Tao Yu},
      year={2024},
      eprint={2404.07972},
      archivePrefix={arXiv},
      primaryClass={cs.AI}
}

@inproceedings{sparse,
  title     = {Selective Learning for Sample-Efficient Training in Multi-Agent Sparse Reward Tasks},
  author    = {Chen, Xinning and Liu, Xuan and Ba, Yanwen and Zhang, Shigeng and Ding, Bo and Li, Kenli},
  booktitle = {Proceedings of the Thirty-Third International Joint Conference on
               Artificial Intelligence, {IJCAI-24}},
  publisher = {International Joint Conferences on Artificial Intelligence Organization},
  editor    = {Kate Larson},
  pages     = {8384--8388},
  year      = {2024},
  month     = {8},
  note      = {Sister Conferences Best Papers},
  doi       = {10.24963/ijcai.2024/927},
  url       = {https://doi.org/10.24963/ijcai.2024/927},
}

@misc{li2025saltstepleveladvantageassignment,
      title={SALT: Step-level Advantage Assignment for Long-horizon Agents via Trajectory Graph}, 
      author={Jiazheng Li and Yawei Wang and David Yan and Yijun Tian and Zhichao Xu and Huan Song and Panpan Xu and Lin Lee Cheong},
      year={2025},
      eprint={2510.20022},
      archivePrefix={arXiv},
      primaryClass={cs.LG},
      url={https://arxiv.org/abs/2510.20022}, 
}

@misc{liu2025agenticreinforcementlearningimplicit,
      title={Agentic Reinforcement Learning with Implicit Step Rewards}, 
      author={Xiaoqian Liu and Ke Wang and Yuchuan Wu and Fei Huang and Yongbin Li and Junge Zhang and Jianbin Jiao},
      year={2025},
      eprint={2509.19199},
      archivePrefix={arXiv},
      primaryClass={cs.CL},
      url={https://arxiv.org/abs/2509.19199}, 
}

@misc{dreamerv3,
      title={Mastering Diverse Domains through World Models}, 
      author={Danijar Hafner and Jurgis Pasukonis and Jimmy Ba and Timothy Lillicrap},
      year={2024},
      eprint={2301.04104},
      archivePrefix={arXiv},
      primaryClass={cs.AI},
      url={https://arxiv.org/abs/2301.04104}, 
}

@article{nq,
    title = "Natural Questions: A Benchmark for Question Answering Research",
    author = "Kwiatkowski, Tom  and
      Palomaki, Jennimaria  and
      Redfield, Olivia  and
      Collins, Michael  and
      Parikh, Ankur  and
      Alberti, Chris  and
      Epstein, Danielle  and
      Polosukhin, Illia  and
      Devlin, Jacob  and
      Lee, Kenton  and
      Toutanova, Kristina  and
      Jones, Llion  and
      Kelcey, Matthew  and
      Chang, Ming-Wei  and
      Dai, Andrew M.  and
      Uszkoreit, Jakob  and
      Le, Quoc  and
      Petrov, Slav",
    editor = "Lee, Lillian  and
      Johnson, Mark  and
      Roark, Brian  and
      Nenkova, Ani",
    journal = "Transactions of the Association for Computational Linguistics",
    volume = "7",
    year = "2019",
    address = "Cambridge, MA",
    publisher = "MIT Press",
    url = "https://aclanthology.org/Q19-1026/",
    doi = "10.1162/tacl_a_00276",
    pages = "452--466",
}

@inproceedings{triviaqa,
    title = "{T}rivia{QA}: A Large Scale Distantly Supervised Challenge Dataset for Reading Comprehension",
    author = "Joshi, Mandar  and
      Choi, Eunsol  and
      Weld, Daniel  and
      Zettlemoyer, Luke",
    editor = "Barzilay, Regina  and
      Kan, Min-Yen",
    booktitle = "Proceedings of the 55th Annual Meeting of the Association for Computational Linguistics (Volume 1: Long Papers)",
    month = jul,
    year = "2017",
    address = "Vancouver, Canada",
    publisher = "Association for Computational Linguistics",
    url = "https://aclanthology.org/P17-1147/",
    doi = "10.18653/v1/P17-1147",
    pages = "1601--1611"
}

@inproceedings{popqa,
    title = "When Not to Trust Language Models: Investigating Effectiveness of Parametric and Non-Parametric Memories",
    author = "Mallen, Alex  and
      Asai, Akari  and
      Zhong, Victor  and
      Das, Rajarshi  and
      Khashabi, Daniel  and
      Hajishirzi, Hannaneh",
    editor = "Rogers, Anna  and
      Boyd-Graber, Jordan  and
      Okazaki, Naoaki",
    booktitle = "Proceedings of the 61st Annual Meeting of the Association for Computational Linguistics (Volume 1: Long Papers)",
    month = jul,
    year = "2023",
    address = "Toronto, Canada",
    publisher = "Association for Computational Linguistics",
    url = "https://aclanthology.org/2023.acl-long.546/",
    doi = "10.18653/v1/2023.acl-long.546",
    pages = "9802--9822"
}

@inproceedings{yang2018hotpotqa,
  title={{HotpotQA}: A Dataset for Diverse, Explainable Multi-hop Question Answering},
  author={Yang, Zhilin and Qi, Peng and Zhang, Saizheng and Bengio, Yoshua and Cohen, William W. and Salakhutdinov, Ruslan and Manning, Christopher D.},
  booktitle={Conference on Empirical Methods in Natural Language Processing ({EMNLP})},
  year={2018}
}

@inproceedings{xanh2020_2wikimultihop,
    title = "Constructing A Multi-hop {QA} Dataset for Comprehensive Evaluation of Reasoning Steps",
    author = "Ho, Xanh  and
      Duong Nguyen, Anh-Khoa  and
      Sugawara, Saku  and
      Aizawa, Akiko",
    booktitle = "Proceedings of the 28th International Conference on Computational Linguistics",
    month = dec,
    year = "2020",
    address = "Barcelona, Spain (Online)",
    publisher = "International Committee on Computational Linguistics",
    url = "https://www.aclweb.org/anthology/2020.coling-main.580",
    pages = "6609--6625",
}

@article{trivedi-etal-2022-musique,
    title = "{M}u{S}i{Q}ue: Multihop Questions via Single-hop Question Composition",
    author = "Trivedi, Harsh  and
      Balasubramanian, Niranjan  and
      Khot, Tushar  and
      Sabharwal, Ashish",
    editor = "Roark, Brian  and
      Nenkova, Ani",
    journal = "Transactions of the Association for Computational Linguistics",
    volume = "10",
    year = "2022",
    address = "Cambridge, MA",
    publisher = "MIT Press",
    url = "https://aclanthology.org/2022.tacl-1.31/",
    doi = "10.1162/tacl_a_00475",
    pages = "539--554"
}

@inproceedings{bamboogle,
    title = "Measuring and Narrowing the Compositionality Gap in Language Models",
    author = "Press, Ofir  and
      Zhang, Muru  and
      Min, Sewon  and
      Schmidt, Ludwig  and
      Smith, Noah  and
      Lewis, Mike",
    editor = "Bouamor, Houda  and
      Pino, Juan  and
      Bali, Kalika",
    booktitle = "Findings of the Association for Computational Linguistics: EMNLP 2023",
    month = dec,
    year = "2023",
    address = "Singapore",
    publisher = "Association for Computational Linguistics",
    url = "https://aclanthology.org/2023.findings-emnlp.378/",
    doi = "10.18653/v1/2023.findings-emnlp.378",
    pages = "5687--5711"
}

@misc{agentsurvey1,
      title={Large Language Model Agent: A Survey on Methodology, Applications and Challenges}, 
      author={Junyu Luo and Weizhi Zhang and Ye Yuan and Yusheng Zhao and Junwei Yang and Yiyang Gu and Bohan Wu and Binqi Chen and Ziyue Qiao and Qingqing Long and Rongcheng Tu and Xiao Luo and Wei Ju and Zhiping Xiao and Yifan Wang and Meng Xiao and Chenwu Liu and Jingyang Yuan and Shichang Zhang and Yiqiao Jin and Fan Zhang and Xian Wu and Hanqing Zhao and Dacheng Tao and Philip S. Yu and Ming Zhang},
      year={2025},
      eprint={2503.21460},
      archivePrefix={arXiv},
      primaryClass={cs.CL},
      url={https://arxiv.org/abs/2503.21460}, 
}

\newpage
\appendix
\section{Implementation Details}
\label{appendix:prompts}

This appendix provides the implementation details in \ourmethod{} training.

\subsection{State Prediction Prompt Template}

The state prediction task prompts the model with the current observation, an action, and historical context, asking it to predict the resulting next observation. The template structure is:

\begin{small}
\begin{verbatim}
You are an expert agent operating in an interactive environment.

Your task is: [TASK_DESCRIPTION]

Prior to this step, you have already taken N step(s).
Below are the history observations and the corresponding
actions you took:
[Observation 1: "...", Action 1: "..."]
[Observation 2: "...", Action 2: "..."]

You are now at step [STEP_NUMBER] and your current observation is:
[CURRENT_OBSERVATION]

You take the action: [ACTION].

Please predict the observation after taking this action.
Present your prediction within <observation> </observation> tags.
\end{verbatim}
\end{small}

The expected response format is: 

\texttt{<observation>predicted next observation</observation>}

\subsection{Inverse Dynamics Prompt Template}

The inverse dynamics task prompts the model with two consecutive observations and a list of admissible actions, asking it to infer which action caused the transition. The template structure is:

\begin{small}
\begin{verbatim}
You are an expert agent operating in an interactive environment.

Your task is: [TASK_DESCRIPTION]

Prior to this step, you have already taken N step(s).
Below are the history observations and the corresponding
actions you took:
[Observation 1: "...", Action 1: "..."]
[Observation 2: "...", Action 2: "..."]

You are now at step [STEP_NUMBER].
Your current observation is: [CURRENT_OBSERVATION]
In the next timestep, the environment observation becomes:
[NEXT_OBSERVATION]

You must select exactly one of the following admissible
actions to explain the transition:
- action_1
- action_2
- action_3
...
Choose the single admissible action that best explains this transition.
Present your chosen action within <action> </action> tags.
\end{verbatim}
\end{small}

The expected response format is: 

\texttt{<action>predicted action</action>}

\subsection{Details in Prompt Construction}

\paragraph{History Window}
In both prompt templates, we use a sliding window of the most recent 2 observation-action pairs to provide context while maintaining a manageable prompt length. This balances the need for temporal context with computational efficiency, and is consistent with the default configuration in the verl-agent framework~\cite{gigpo}.

\paragraph{Action Shuffling}
For the inverse dynamics task, admissible actions are randomly shuffled before being presented in the prompt to prevent the model from exploiting positional biases. The ground truth action is always included in the admissible set to ensure the task is solvable.

\paragraph{Structured Output}
Both tasks require the model to output predictions within XML-style tags (\texttt{<observation>} and \texttt{<action>}). This structured format enables reliable extraction of predictions for automated evaluation.

\section{SearchQA Experiments}
\label{appendix:searchqa}
We also evaluate the multi-turn tool calling performance of \ourmethod{} on search-augmented QA tasks, including single-hop QA datasets (NQ~\cite{nq}, TriviaQA~\cite{triviaqa}, and PopQA~\cite{popqa}) and multi-hop QA datasets (HotpotQA~\cite{yang2018hotpotqa} , 2Wiki~\cite{xanh2020_2wikimultihop}, MuSiQue~\cite{trivedi-etal-2022-musique}, and Bamboogle~\cite{bamboogle}). Following the trainingsettings in verl-agent~\cite{gigpo}, we employ Qwen2.5-3B-Instruct as the backbone, train the model on NQ and HotpotQA datasets, and test on all 7 benchmarks. As illustrated in Table~\ref{tab:qa_benchmarks}, our method yields consistent improvements across all seven benchmarks compared to the GiGPO baseline, achieving a +1.88\% gain in average performance. This enhancement suggests that internalizing environment dynamics remains beneficial even in more stochastic search-based environments.

\begin{table}[htbp]
    \centering
    \caption{Performance on SearchQA benchmarks.}
    \label{tab:qa_benchmarks}
    \resizebox{\linewidth}{!}{
    \begin{tabular}{lcccccccc}
        \toprule
        \textbf{Method} & \textbf{NQ} & \textbf{TriviaQA} & \textbf{PopQA} & \textbf{HotpotQA} & \textbf{2Wiki} & \textbf{MuSiQue} & \textbf{Bamboogle} & \textbf{Avg.} \\
        \midrule
        GiGPO           & 42.69 & 60.41 & 42.13 & 37.94 & 37.32 & 13.07 & 64.52 & 43.33 \\
        \ourmethod{}-GiGPO      & \textbf{44.55} & \textbf{62.07} & \textbf{45.82} & \textbf{38.32} & \textbf{39.66} & \textbf{13.73} & \textbf{65.32} & \textbf{45.21} \\
        \bottomrule
    \end{tabular}}
\end{table}

\section{Limitations and Future Work}
\label{append:limitations}
Despite its effectiveness, \ourmethod{} has several limitations. First, the current state-prediction objective is designed for LLMs. For multi-modal models such as VLM, this objective remains to be more specific designed. Second, despite the insensitivity of the initial auxiliary weights $\lambda_{SP}$ and $\lambda_{ID}$, more advanced mechanisms such as self-adaptive tuning should be applied to balance these learning objectives automatically. Finally, the supervision signals from environment dynamics explored in \ourmethod{} are not exhaustive. There should exist various other mechanisms that remain to be investigated. As the comprehensive explorations discussed above are beyond the scope of this paper, we leave them for future work.


\end{document}